%% file: main.tex
\documentclass[sigconf]{acmart}

\copyrightyear{2027}
\acmYear{2027}
\setcopyright{cc}
\setcctype{by} 
\acmConference[WSDM 2027]{Proceedings of the Twentieth ACM International Conference on Web Search and Data Mining}{February 15--19, 2027}{Hong Kong, Hong Kong.}
\acmBooktitle{Proceedings of the Twentieth ACM International Conference on Web Search and Data Mining (WSDM 2027), February 15--19, 2027, Hong Kong, Hong Kong}
\acmISBN{979-8-4007-3117-4/2027/02}
\acmDOI{10.1145/3862322.3863518}

\input{utils/general_utils}
\input{utils/includes}

\input{utils/math_utils}

\begin{document}

\title[Large Language Model Unlearning Under the Microscope]{Large Language Model Unlearning Under the Microscope: A Benchmark Study}

\author{Chongyu Fan}
\authornote{Equal contribution.}
\orcid{0009-0008-4228-8942}
\affiliation{%
  \institution{Michigan State University}
  \city{East Lansing}
  \state{Michigan}
  \country{USA}
}
\email{fanchon2@msu.edu}

\author{Changsheng Wang}
\authornotemark[\value{footnote}]
\orcid{0009-0007-0957-638X}
\affiliation{%
  \institution{Michigan State University}
  \city{East Lansing}
  \state{Michigan}
  \country{USA}
}
\email{wangc168@msu.edu}

\author{Yancheng Huang}
\authornote{Equal contribution.}
\orcid{0000-0001-8271-6085}
\affiliation{%
  \institution{Michigan State University}
  \city{East Lansing}
  \state{Michigan}
  \country{USA}
}
\email{huang341@msu.edu}

\author{Soumyadeep Pal}
\authornotemark[\value{footnote}]
\orcid{0000-0003-0527-1224}
\affiliation{%
  \institution{Michigan State University}
  \city{East Lansing}
  \state{Michigan}
  \country{USA}
}
\email{palsoum1@msu.edu}

\author{Sijia Liu}
\orcid{0000-0003-2817-6991}
\affiliation{%
  \institution{Michigan State University}
  \city{East Lansing}
  \state{Michigan}
  \country{USA}
}
\affiliation{%
  \institution{MIT-IBM Watson AI Lab}
  \city{Cambridge}
  \state{Massachusetts}
  \country{USA}
}
\email{liusiji5@msu.edu}

\renewcommand{\shortauthors}{Chongyu Fan, Changsheng Wang, Yancheng Huang, Soumyadeep Pal, and Sijia Liu}

\input{secs/abstract}

\ccsdesc[500]{Computing methodologies~Natural language processing}
\ccsdesc[500]{Computing methodologies~Machine learning}
\ccsdesc[300]{Security and privacy~Domain-specific security and privacy architectures}

\keywords{machine unlearning; large language models; benchmark evaluation; adversarial robustness}

\maketitle
\setlength{\textfloatsep}{6.5pt plus 1.5pt minus 1pt}
\setlength{\dbltextfloatsep}{7.5pt plus 1.5pt minus 1pt}
\setlength{\floatsep}{6pt plus 1.5pt minus 1pt}
\setlength{\dblfloatsep}{6.5pt plus 1.5pt minus 1pt}
\setlength{\intextsep}{6.5pt plus 1.5pt minus 1pt}

\input{secs/introduction}
\input{secs/related_work}
\input{secs/unlearning_method_SLiu}
\input{secs/experimental_setup}
\input{secs/sec_eval_no_rob_SLiu}
\input{secs/sec_rob_SLiu}

\input{secs/conclusion}

\input{secs/limitations}
\input{secs/ai_usage}

\begin{acks}
This work was supported in part by  the U.S. National Science Foundation (NSF) under CISE Core Award
IIS-2504263 and CAREER Award IIS-2338068, the Coefficient Giving AI Safety Research
Award, and the Schmidt Sciences Trustworthy AI Award.
\end{acks}

\FloatBarrier
\clearpage
\input{secs/impact}

\bibliographystyle{ACM-Reference-Format}
\balance
\bibliography{refs/MU,refs/MU_SLiu,refs/Smooth,refs/MU_CF}

\end{document}

%% file: utils/general_utils.tex
\usepackage{pifont}
\newcommand{\cmark}{\textcolor{green}{\ding{51}}}
\newcommand{\xmark}{\textcolor{red}{\ding{55}}}

\usepackage{colortbl}
\definecolor{Gray}{gray}{0.93}
\definecolor{Orange}{rgb}{1,0.5,0}
\definecolor{DGray}{gray}{0.83}
\definecolor{LightCyan}{rgb}{0.88,1,1}

\definecolor{mr}{RGB}{251, 111, 111}

\definecolor{SL_color}{rgb}{0.858, 0.188, 0.478}

\definecolor{CW_color}{rgb}{0.607, 0.349, 0.714}

\definecolor{lightgreen}{HTML}{D9EAD3} 
\definecolor{lightred}{HTML}{F4CCCC}   

\definecolor{original}{HTML}{D96C5F}   
\definecolor{CRej}{HTML}{7CAAD9}       
\definecolor{CRep}{HTML}{F2B15C}     
\definecolor{CDiv}{HTML}{3C9A87}      
\definecolor{MCQ}{HTML}{800080}    
\definecolor{Open-QA}{HTML}{808080}    

\newcommand{\MDiv}{\textcolor{CDiv}{\textit{divergence-driven optimization}}}

\newcommand{\MRep}{\textcolor{CRep}{\textit{representation misalignment}}}

\newcommand{\MRej}{\textcolor{CRej}{\textit{rejection-based targeted unlearning}}}

\newcommand{\MDivB}{{divergence-driven optimization}}

\newcommand{\MRepB}{{representation misalignment}}

\newcommand{\MRejB}{rejection-based targeted unlearning}

%% file: utils/includes.tex
\usepackage{subfigure}
\usepackage{pifont}
\usepackage{colortbl}
\usepackage{multirow}
\usepackage{mathtools}
\usepackage[capitalize,noabbrev]{cleveref}
\usepackage[most]{tcolorbox}
\usepackage{adjustbox}
\usepackage{placeins}
\usepackage{balance}

\makeatletter
\@ifundefined{KV@Gin@artifact}{\define@key{Gin}{artifact}[]{}}{\relax}
\makeatother
\expandafter\def\expandafter\UrlBreaks\expandafter{\UrlBreaks\do-\do_}

\theoremstyle{plain}

\theoremstyle{definition}

\theoremstyle{remark}

%% file: utils/math_utils.tex
\usepackage{bm}

\def\eqref#1{(\ref{#1})}

\def\1{\bm{1}}

\newcommand{\Dr}{\mathcal{D_{\mathrm{r}}}}
\newcommand{\Df}{\mathcal{D_{\mathrm{f}}}}

\DeclareMathAlphabet{\mathsfit}{\encodingdefault}{\sfdefault}{m}{sl}
\SetMathAlphabet{\mathsfit}{bold}{\encodingdefault}{\sfdefault}{bx}{n}

\DeclareMathOperator*{\minimize}{\text{minimize}}

\newcommand{\btheta}{{\boldsymbol{\theta}}}

\newcommand{\ellf}{{\ell_{\mathrm{f}}}}
\newcommand{\ellr}{{\ell_{\mathrm{r}}}}

%% file: secs/abstract.tex
\begin{abstract}
Machine unlearning for large language models (LLMs), known as LLM unlearning, aims to remove \textit{undesired} data, knowledge, and behaviors 
while preserving model capabilities. Despite rapid progress over the past two years, research in LLM unlearning remains fragmented, with limited clarity on what constitutes effective unlearning and how it should be rigorously evaluated. In this work, we present a principled taxonomy of \textit{twelve} recent stateful unlearning methods, grouped into three methodological families: {\MDiv}, {\MRep}, and {\MRej}. Building on this taxonomy, we revisit the evaluation of unlearning effectiveness (UE), utility retention (UT), and robustness (Rob), focusing on WMDP for hazardous knowledge and associated model capability removal.
Our analysis shows that current evaluations, dominated by multiple-choice question (MCQ) accuracy, offer only a narrow perspective, often overstating success while overlooking the model’s generation behavior. To address this gap, we introduce open question answering (Open-QA) metrics to better capture generative performance and reveal the strengths and limitations of different unlearning method families. Furthermore, we demonstrate that robustness requires finer-grained analysis. For example, vulnerabilities differ substantially between \textit{in-domain relearning} and \textit{out-of-domain fine-tuning}, even though both fall under model-level attacks. Through this study, we provide a full-stack re-examination of LLM unlearning and actionable guidance for designing and evaluating future methods. Code and model checkpoints are publicly available on \href{https://github.com/OPTML-Group/Unlearn-FullStack}{GitHub} and \href{https://huggingface.co/collections/OPTML-Group/unlearn-fullstack}{Hugging Face}.
\end{abstract}

%% file: secs/introduction.tex
\section{Introduction}

With the rapid advancement of large language models (LLMs), their tendency to memorize and regurgitate training data and knowledge has raised serious concerns regarding privacy, safety, and property \citep{liu2024towards,che2025model,barez2025open}. From a generative behavior perspective, this risk appears as producing sensitive information, harmful content, or copyrighted material, motivating the urgent need for \emph{machine unlearning}, which selectively removes undesired data, knowledge, or behaviors while preserving general downstream utility \citep{yao2024large,maini2024tofu,liu2024rethinking,wang2025reasoning}.

The growing interest in \textit{LLM unlearning} has spurred the development of a broad range of algorithms. In this work, we study \textbf{twelve} representative methods and organize them into a \textit{taxonomy} with three families: {\MDiv}, {\MRep}, and {\MRej}.
A summary is provided in \textbf{Table\,\ref{tab:taxonomy}}, where each method is associated with one or more of six benchmarks: WMDP \citep{li2024wmdp}, MUSE \citep{shi2024muse}, TOFU \citep{maini2024tofu}, WHP \citep{eldan2023whos}, PKU-SafeRLHF \citep{ji2024pku}, and Circuit Breakers \citep{zou2024improving}.
Specifically, divergence-driven optimization drives the model away from a reference distribution \citep{yao2024large,fan2025towards,wang2025invariance}, with negative preference optimization (NPO) \citep{zhang2024negative} and simple NPO (SimNPO) \citep{fan2024simplicity} as examples. Representation misalignment disrupts the embeddings of forget data relative to their reference model representations \citep{zou2024improving,tamirisa2024tamper,sheshadri2024latent}, such as representation misdirection for unlearning (RMU) \citep{li2024wmdp}. Rejection-based methods enforce unlearning by producing explicit rejections to forget queries \citep{yuan2024closer,singh2025unlearning}, with I Don’t Know (IDK) \citep{maini2024tofu} as a representative case. 

\input{tabs/unlearn_method}

As shown in Table~\ref{tab:taxonomy}, existing LLM unlearning approaches are evaluated on off-the-shelf benchmarks that vary both in methodological assumptions and evaluation criteria. The most popular choice is the Weapons of Mass Destruction Proxy (WMDP) benchmark \citep{li2024wmdp}, which is valued for its practical focus on erasing harmful knowledge and generation behaviors, \textit{i.e.}, \textit{capability removal}. This stands in contrast to benchmarks for \textit{data- or instance-wise unlearning}, which are mostly motivated by privacy considerations and typically rely on  {additional fine-tuning on the unlearn dataset} to first induce memorization of the target content before unlearning is studied, as in TOFU \citep{maini2024tofu} and MUSE \citep{shi2024muse}.
Although this additional fine-tuning serves as the starting point for unlearning, it degrades the model's general utility due to overshooting on the unlearning domain.
Therefore, given the strong practical relevance of WMDP (focusing on the removal of undesired knowledge and capabilities without introducing any pre-unlearning fine-tuning), we adopt \textit{WMDP as the primary benchmark in this work}.
In WMDP, unlearning performance is assessed along two dimensions: \emph{unlearning effectiveness} (UE) and \emph{utility retention} (UT). However, these evaluations mostly rely on multiple-choice questions (MCQ), where the model selects from predefined options \citep{li2024wmdp}. Such evaluations risk \textit{obscuring} the model’s free-form generation after unlearning, limiting UE and UT assessment beyond predefined options.

In addition to UE and UT, \emph{unlearning robustness} (Rob) is critical, as forgotten knowledge can re-emerge once an unlearned model is ``attacked'' \citep{hu2024jogging,lucki2024adversarial,tamirisa2024tamper,che2025model,lynch2024eight}. Such attacks can take various forms, ranging from model-level perturbations \citep{hu2024jogging,tamirisa2024tamper,fan2025towards,wang2025invariance,zhang2024catastrophic} to input-level manipulations \citep{lucki2024adversarial,lynch2024eight}. However, systematic studies of unlearning robustness and its relationship to the unlearning method taxonomy (\textit{e.g.}, with and without robust design) and to different attack settings remain limited.


Given the proposed taxonomy of unlearning methods and the complexities of unlearning evaluations, the key research question we aim to address is:
\begin{tcolorbox}[before skip=2.8mm, after skip=2.8mm, boxsep=0.0cm, middle=0.0cm, top=1.4mm, bottom=1.4mm, left=2.0mm, right=2.0mm, boxrule=0.6pt]
\centering
\textit{\textbf{(Q)} Can we conduct a full-stack investigation of LLM unlearning that yields methodology-wise insights across all key metrics (UE, UT, and Rob)?}
\end{tcolorbox}

To tackle (Q), we first draw methodological insights from our taxonomy: {\MDiv}, {\MRep}, and {\MRej}. We then revisit the conventional unlearning assessments, UE and UT, and argue that they should also be evaluated through open question answering (Open-QA), where the model generates free-form responses, beyond MCQ. 
For instance, MCQ-based MMLU assessments of UT \textit{cannot} fully capture the over-forgetting issue in {\MDivB} methods such as NPO \citep{zhang2024negative}, nor can MCQ-based UE evaluations fully reflect the performance of {\MRejB}. Furthermore, in the dimension of Rob, we find that robustness in LLM unlearning should be examined at a finer granularity, including (i) in-domain relearning \citep{hu2024jogging,fan2025towards}, where the model is fine-tuned on a subset of forget data, and (ii) out-of-domain fine-tuning \citep{wang2025invariance}, where the model is adapted to unrelated downstream tasks. Divergence-driven optimization  methods are generally more resilient to in-domain relearning, whereas {\MRepB} methods show stronger resistance to out-of-domain fine-tuning. In addition, robustness against input-level attacks such as jailbreaking is more closely aligned with in-domain relearning.

Although prior work has made initial attempts to study LLM unlearning \textit{systematically} \citep{dorna2025openunlearning,wang2024towards,che2025model,feng2025existing}, our work is not merely a re-benchmarking exercise. Instead, our contributions lie in a methodology-driven, taxonomy-guided analysis of previously overlooked aspects of unlearning: from categorizing method families, to rethinking UE/UT evaluation beyond MCQ, to analyzing multi-faceted robustness through the lens of these methodological principles.
We summarize \textbf{our contributions} below.

\ding{182} We establish a taxonomy of existing unlearning methods, categorizing them into {\MDiv}, {\MRep}, and {\MRej}. \ding{183} We revisit evaluation practices beyond MCQ to incorporate Open-QA metrics, which better capture generative performance and reveal fundamental characteristics of different method families across UT and UE. \ding{184} We revisit the robustness of LLM unlearning under model-level attacks (in-domain relearning, out-of-domain fine-tuning, and quantization) and input-level jailbreak attacks, showing how these robustness dimensions are connected.




%% file: tabs/unlearn_method.tex
\begin{table}[thb]
\centering
\caption{\small Taxonomy of twelve unlearning methods grouped into {\MDiv}, {\MRep}, and {\MRej}. A checkmark (\cmark) marks incorporation of robust unlearning designs. Benchmarks include WMDP (W), MUSE (M), TOFU (T), WHP (H), PKU-SafeRLHF (P), and Circuit Breakers (C), evaluated by unlearning effectiveness (UE) and utility retention (UT) under MCQ or Open-QA, and robustness (Rob) at the model or input level.}
\label{tab:taxonomy}
\renewcommand{\arraystretch}{1.05}
\resizebox{1.0\linewidth}{!}{%
\begin{tabular}{c|c|c|cc|cc|cc}
\toprule[1pt]
\multirow{2}{*}{\textbf{Method}} & \multirow{2}{*}{\textbf{Reference}} & \multirow{2}{*}{\textbf{\begin{tabular}[c]{@{}c@{}}Robust\\Design\end{tabular}}} & \multicolumn{2}{c|}{\textbf{UE}} & \multicolumn{2}{c|}{\textbf{UT}} & \multicolumn{2}{c}{\textbf{Rob}} \\
\cline{4-9}
& & & \textbf{MCQ} & \textbf{Open-QA} & \textbf{MCQ} & \textbf{Open-QA} & \textbf{Model} & \textbf{Input} \\
\midrule
\rowcolor{LightCyan!50}
\multicolumn{9}{c}{\textbf{\textcolor{CDiv}{\textit{Divergence-driven optimization}}}} \\
\midrule
\rowcolor{gray!20}
\textbf{GradDiff} & \citep{yao2024large} & \xmark & & P & & P & & \\
\textbf{NPO} & \citep{zhang2024negative} & \xmark & & T & & T & & \\
\rowcolor{gray!20}
\textbf{SimNPO} & \citep{fan2024simplicity} & \xmark & W & T/M & W & T/M & T & \\
\textbf{NPO+SAM} & \citep{fan2025towards} & \cmark & W & M & W & M & W/M & W \\
\rowcolor{gray!20}
\textbf{NPO+IRM} & \citep{wang2025invariance} & \cmark & W & M & W & M & W/M & \\
\midrule
\rowcolor{LightCyan!50}
\multicolumn{9}{c}{\textbf{\textcolor{CRep}{\textit{Representation misalignment}}}} \\
\midrule
\textbf{RMU} & \citep{li2024wmdp} & \xmark & W & & W & W & & W \\
\rowcolor{gray!20}
\textbf{RR} & \citep{zou2024improving} & \xmark & C & & C & C & & C \\
\textbf{RMU+LAT} & \citep{sheshadri2024latent} & \cmark & W/H & & W/H & W & W & H \\
\rowcolor{gray!20}
\textbf{TAR} & \citep{tamirisa2024tamper} & \cmark & W & & W & & W & \\
\midrule
\rowcolor{LightCyan!50}
\multicolumn{9}{c}{\textbf{\textcolor{CRej}{\textit{Rejection-based targeted unlearning}}}} \\
\midrule
\textbf{ELM} & \citep{gandikota2024erasing} & \xmark & W/H & & W/H & W/H & & W \\
\rowcolor{gray!20}
\textbf{DPO} & \citep{zhang2024negative} & \xmark & & T & T & T & & \\
\textbf{IDK+AP} & \citep{yuan2024closer} & \xmark & & T & T & T & & \\


\bottomrule[1pt]
\end{tabular}
}
\end{table}

%% file: secs/related_work.tex
\section{Related Work}

\noindent \textbf{LLM unlearning and benchmarking studies.} Recent work on LLM unlearning~\citep{liu2024rethinking,maini2024tofu,liu2024large,yao2024large} has made progress in mitigating risks such as copyright infringement~\citep{eldan2023whos}, privacy leakage~\citep{hu2024jogging,wu2023depn}, and harmful content generation~\citep{li2024wmdp,lu2022quark}; see Sec.\,\ref{sec:taxonomy} for detailed introductions to these method families. 

Several benchmarks have been proposed to evaluate LLM unlearning~\citep{li2024wmdp,maini2024tofu,jin2024rwku,shi2024muse,eldan2023whos,blanco2025digital}. Some construct unlearning targets by fine-tuning on a domain corpus, such as WHP~\citep{eldan2023whos}, TOFU~\citep{maini2024tofu}, and MUSE~\citep{shi2024muse}. 
Others avoid such fine-tuning and better reflect real-world settings, with WMDP~\citep{li2024wmdp} as a prime example.

There is also growing interest in unifying evaluation practices for LLM unlearning and formalizing desiderata such as faithfulness and robustness. For example, Wang et al.~\citep{wang2024towards} propose the UWC (Unlearning with Control) evaluation framework and identify Extraction Strength as a core metric for fair comparison and calibration across unlearning methods. Dorna et al.~\citep{dorna2025openunlearning} introduce OpenUnlearning, a standardized platform that integrates multiple unlearning methods and metrics and assesses metric faithfulness and robustness through stress tests. However, both primarily focus on TOFU for data-wise unlearning.

Beyond the above efforts, related work has examined LLM unlearning from several complementary perspectives. From the viewpoint of optimization and objective design, several studies analyze the unlearning–retention trade-off \citep{wang2025gru,wang2025rethinking}. From the perspective of unlearning evaluation, Feng et al.~\citep{feng2025existing} identify principles such as minimal information injection and downstream task awareness for evaluation design.
Furthermore, adaptive test generation can expose hidden knowledge gaps \citep{ko2025probing}, and privacy risk has been shown to be not fully captured by verbatim memorization proxies \citep{sanderrethinking}.

Unlike existing benchmarks and platforms, which primarily serve as testbeds or standardize evaluation pipelines, our work provides a taxonomy-guided, method-level analysis of LLM unlearning. We introduce a principled categorization of unlearning methods and use it to systematically analyze (i) MCQ versus Open-QA evaluation, showing that MCQ often overstates unlearning success and obscures over-forgetting and utility loss, and (ii) robustness across both model-level and input-level attacks, including finer-grained weight perturbations (in-domain vs. out-of-domain) and their connections to input-level robustness.

\noindent \textbf{Adversarial robustness of LLM unlearning.} Robustness has emerged as a central challenge for unlearning, as unlearned models remain vulnerable to diverse attacks~\citep{lucki2024adversarial,che2025model,hu2024jogging}. Model-level attacks can restore forgotten content through light fine-tuning~\citep{lucki2024adversarial,hu2024jogging,qi2023fine,halawi2024covert,lermen2023lora,huang2024harmful}, pruning~\citep{jain2023mechanistically,wei2024assessing,lee2018snip}, or quantization~\citep{zhang2024catastrophic}, which may re-expose knowledge by compressing weights. Representation-level attacks perturb embeddings or hidden activations to revive residual traces~\citep{schwinn2024soft,sheshadri2024latent}, while input-level attacks exploit adversarial prompting or query optimization~\citep{chao2025jailbreaking,shin2020autoprompt}, ranging from gradient-guided search~\citep{zou2023universal} to perplexity-based strategies~\citep{sadasivan2024fast}. 

In response, several defenses have been proposed. Adversarial training strengthens robustness against such attacks~\citep{sheshadri2024latent}, and meta-learning strategies further enhance defense~\citep{tamirisa2024tamper,sondej2025robust}. Sharpness-aware minimization encourages flat minima to reduce susceptibility to relearning~\citep{fan2025towards}. Invariance-based regularization introduces robustness via invariant risk principles~\citep{wang2025invariance}, while distillation transfers knowledge into partially noised student models~\citep{lee2025distillation}.  

From the perspective of robustness evaluation, Che et al.~\citep{che2025model} benchmark the robustness of LLM unlearning against a wide range of post-unlearning weight perturbations, while Hu et al.~\citep{hu2025blur} examine robustness to forget--retain overlap. However, these studies remain limited: Che et al.~\citep{che2025model} evaluate robustness only using MCQ accuracy, and Hu et al.~\citep{hu2025blur} consider only in-domain relearning, leaving a more comprehensive assessment unexplored.

%% file: secs/unlearning_method_SLiu.tex
\section{A Taxonomy of Unlearning Methods: Design Principles and Insights}
\label{sec:taxonomy}


\noindent \textbf{Problem setup for LLM unlearning.}
The unlearning problem is defined with respect to a dataset whose influence must be erased, denoted as the \textit{forget set} ($\Df$). To preserve model utility, one also specifies a complementary \textit{retain set} ($\Dr$) whose knowledge should remain unaffected. These two sets capture the dual objectives of unlearning: removing information from $\Df$ while maintaining the model's general utility as reflected by $\Dr$. This trade-off can be formulated as a regularized optimization problem \citep{liu2024rethinking}:
\begin{equation}
 \minimize_{\boldsymbol{\theta}} \; \ellf (\boldsymbol{\theta}; {\Df})  
 + \lambda  \ellr(\boldsymbol{\theta}; {\Dr}) ,
\label{eq:prob_LLM_MU}
\end{equation}%
where $\btheta$ are the model parameters to be updated, $\lambda \geq 0$ is a regularization weight balancing forgetting and retention, and $\ellf$ and $\ellr$ denote loss terms for the forget and retain objectives, respectively.

A large body of work has investigated different design choices for instantiating~\eqref{eq:prob_LLM_MU}.
Based on their methodological principles, we categorize existing approaches into three families: {\MDiv}, {\MRep}, and {\MRej} (summarized in Table\,\ref{tab:taxonomy}).


\noindent \textbf{Divergence-driven optimization for unlearning.} 
The first family of methods designs the forget loss $\ellf$ in~\eqref{eq:prob_LLM_MU} to maximize the ``divergence'' between the prediction logits of the unlearned model and the reference (original) model on the forget set $\Df$. We refer to this class as {\MDiv}, since it explicitly drives the model away from the reference model.  

The most basic instance is gradient ascent (GA) \citep{thudi2022necessity}, which directly increases the prediction loss on $\Df$. However,  GA often pushes the model too far from the reference, leading to collapse \citep{zhang2024negative}. To address this, several GA-type variants have been proposed. Gradient difference (\textbf{GradDiff}) \citep{yao2024large} balances objectives by applying gradient ascent on $\Df$ while using gradient descent on $\Dr$, thereby controlling divergence from the reference.  This approach shares a similar spirit with NegGrad-type methods for classical machine unlearning in discriminative models \citep{kurmanji2023towards}.

Negative preference optimization (\textbf{NPO}) \citep{zhang2024negative} interprets forget samples as negative preferences in the DPO (direct preference optimization) \citep{rafailov2024direct}. This  specifies $\ellf$ with 
\begin{align}
\ell_{\mathrm{NPO}}(\btheta) =  
\mathbb E_{(x,y) \in \Df} \left[
- \tfrac{2}{\beta} \log \sigma \Big( - \beta \log \tfrac{\pi_{\btheta}(y \mid x)}{\pi_{\mathrm{ref}}(y \mid x)} \Big)
\right], \label{eq:NPO_loss}
\end{align}
where $\pi_{\btheta}(y \mid x)$ represents the prediction probability of the model $\btheta$ given
the input-response pair $(x,y)$, and  $\pi_{\mathrm{ref}}$ refers to the reference model.
Simple NPO (\textbf{SimNPO}) \citep{fan2024simplicity} further mitigates reference-model bias in NPO by modifying $\ellf$ to
$- \tfrac{2}{\beta} \log \sigma \!\left( - \tfrac{\beta}{|y|}\log \pi_{\btheta}(y \mid x) \right)$.

Building on NPO, recent works have further addressed robustness gaps in LLM unlearning, particularly sensitivity to model-level attack after unlearning. Examples include \textbf{NPO+SAM} \citep{fan2025towards}, which leverages sharpness-aware minimization (SAM) \citep{foret2021sharpnessaware} to flatten the forget loss landscape, and \textbf{NPO+IRM} \citep{wang2025invariance}, which applies invariant risk minimization (IRM) \citep{arjovsky2019invariant} to encourage robustness.


\noindent \textbf{Representation misalignment for unlearning.}
Another class of methods operates on internal representations, seeking to misalign the representations of $\Df$ with their original representations in the reference model. We term this family {\MRep}. The principle originates from early methods \citep{golatkar2020eternal,chundawat2023can,fan2023salun} that inject randomness into forget data to disrupt memorization and reduce alignment on $\Df$.
The most widely used approach in this family is \textbf{representation misdirection for unlearning (RMU)} \citep{li2024wmdp}, which steers the unlearned model’s hidden states toward a random vector. This formulates the forget loss $\ellf$ as 
\begin{align}
  \ell_{\mathrm{RMU}}(\boldsymbol{\theta}) = 
  \mathbb E_{ x \in \Df} \left[
    || {M_{\boldsymbol{\theta}}(x) - c \cdot {\mathbf u}||}_2^2 
  \right],
  \label{eq:RMU_loss}
\end{align}
where $M_{\boldsymbol{\theta}}$ represents certain intermediate-layer representations of $\btheta$, $c > 0$ is a hyperparameter that controls activation scaling, and $\mathbf{u}$ is a random vector drawn from a standard uniform distribution.


In addition, representation rerouting (\textbf{RR}) \citep{zou2024improving} modifies RMU by replacing the $\ell_2$ norm in $\ell_f$ with the cosine similarity between the unlearned model $M_{\boldsymbol{\theta}}$ and the reference model, in analogy to the classical unlearning method SCRUB \citep{kurmanji2023towards} for discriminative models.
Other approaches, such as tampering attack resistance (\textbf{TAR}) \citep{tamirisa2024tamper} and latent adversarial training (\textbf{LAT}) \citep{sheshadri2024latent}, build on RMU with meta-learning or adversarial training in the latent space to improve unlearning robustness.

\noindent \textbf{Rejection-based unlearning.}
Unlike divergence-driven or representation misalignment approaches, which perform \textit{untargeted} unlearning by discouraging alignment with the forget set, the {\MRej} family enforces \textit{targeted} unlearning through explicit rejection responses to forget queries.
%
A representative method is the I Don’t Know (\textbf{IDK}) strategy \citep{maini2024tofu}, which defines the forget loss $\ellf$ as the prediction loss over rejection-labeled forget data:
\begin{align}
 \ell_{\mathrm{IDK}}(\btheta) =  
 \mathbb E_{(x \in \Df, y \in \mathcal{D}_\text{IDK})} 
 \left [ - \log \pi_{\btheta}(y \mid x) \right ] ,
 \label{eq:IDK_loss}
\end{align}
where $\mathcal{D}_\text{IDK}$ denotes rejection labels in varied formats.

Beyond the simplest IDK formulation, \textbf{DPO} \citep{rafailov2024direct,zhang2024negative} can also be adapted for unlearning by treating rejection as the positive response for forget data. Other extensions include \textbf{IDK+AP} \citep{yuan2024closer}, which, similar in spirit to DPO, introduces an answer preservation (AP) loss that regards the normal response as positive on retain data and the rejection response as positive on forget data, augmenting IDK with an alignment objective.
Similarly, erasing via language modeling (\textbf{ELM}) \citep{gandikota2024erasing} aligns the unlearned model’s outputs with those of a prompted reference model using a predefined prefix (\textit{e.g.}, \textit{``As a novice in bioweapons''}) to steer refusal-like outputs.

\begin{figure}[!htbp]
    \centering
    \includegraphics[width=0.48\columnwidth]{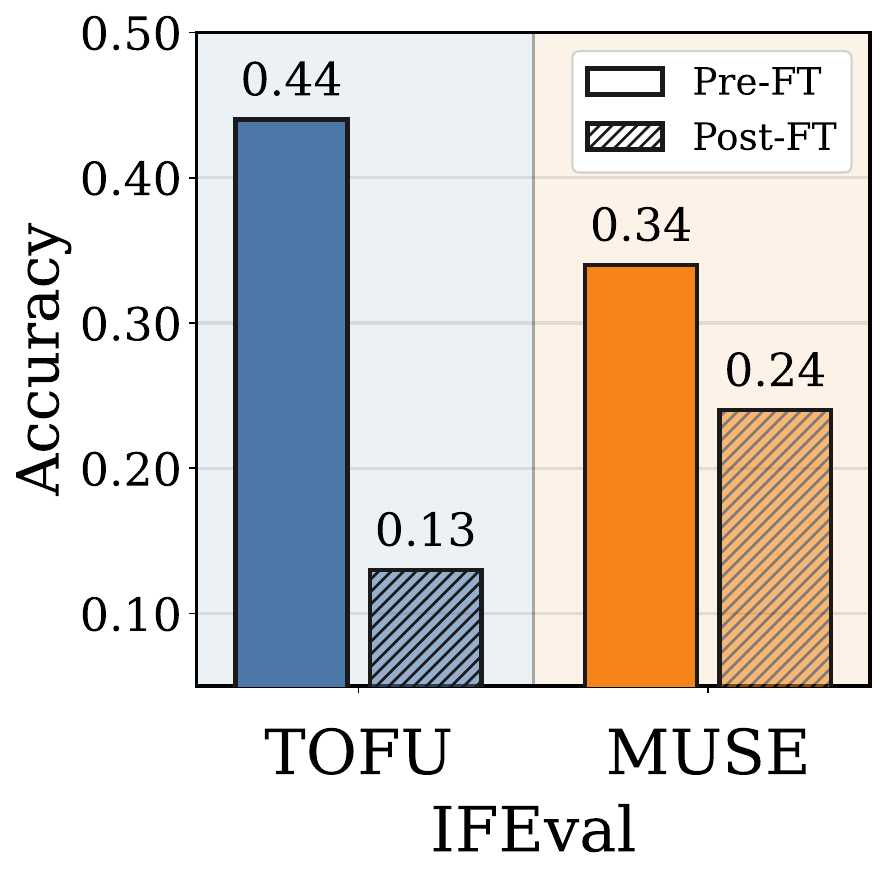}
    \vspace{-1.5mm}
    \caption{\small IFEval accuracy of unlearning-domain pre-fine-tuning (Pre-FT) and post-fine-tuning (Post-FT) models on TOFU and MUSE-News, where Post-FT serves as the starting point for unlearning.}
    \Description{Bar chart comparing IFEval accuracy of pre-fine-tuning and post-fine-tuning models on TOFU and MUSE-News, showing substantial utility drops after domain fine-tuning.}
    \label{fig:tofu_muse}
\end{figure}

\noindent \textbf{Why focus on the WMDP unlearning benchmark?}
The aforementioned methods have been evaluated across different unlearning benchmarks, as summarized in Table~\ref{tab:taxonomy}. However, in this work we primarily focus on \textbf{WMDP}~\citep{li2024wmdp}, as our interest lies in model capability unlearning 
rather than instance-wise unlearning driven by data memorization. Another important reason is that other benchmark datasets require additional fine-tuning on the forget and retain corpora $\mathcal{D}_{\mathrm f} \cup \mathcal{D}_{\mathrm{r}}$, \textit{e.g.}, TOFU~\citep{maini2024tofu} and MUSE~\citep{shi2024muse}, which can already degrade general capabilities and therefore confound utility assessment. In \textbf{Fig.~\ref{fig:tofu_muse}}, we follow the TOFU and MUSE protocols and compare the fine-tuned models (used as the starting point for unlearning) with the pre-fine-tuning models, observing substantial drops in utility; under such settings, analyzing general utility is less informative.

\noindent \textbf{Evaluation metrics.}
Unlearning performance is most commonly evaluated in terms of unlearning effectiveness (\textbf{UE}) and utility retention (\textbf{UT}). 
As shown in Table\,\ref{tab:taxonomy},
these metrics in existing benchmarks can be generally classified into two types: \textit{(i) multiple-choice questions} (\textbf{MCQ}), where the model selects from predefined options, and \textit{(ii) open question-answering} (\textbf{Open-QA}), where the model generates free-form answers. Therefore, we denote $\mathrm{UE}_{\mathrm{MCQ}}$ (or $\mathrm{UT}_{\mathrm{MCQ}}$) and $\mathrm{UE}_{\mathrm{Open\text{-}QA}}$ (or $\mathrm{UT}_{\mathrm{Open\text{-}QA}}$) as the corresponding UE (or UT) metrics, respectively.
However, model capability removal has so far been mostly restricted to MCQ-based evaluation \citep{feng2025existing} rather than Open-QA.
In addition to UE and UT, robustness (\textbf{Rob}) has emerged as another critical dimension in evaluating LLM unlearning. We distinguish two types of robustness assessments: \textit{model-level weight tampering}, like in-domain relearning \citep{hu2024jogging, fan2025towards} and out-of-domain fine-tuning \citep{wang2025invariance}; and \textit{input-level attacks}  \citep{lucki2024adversarial}.
Complete training, evaluation, and robustness configurations are provided in Sec.~\ref{sec:experimental_setup}.

%% file: secs/experimental_setup.tex
\section{Experimental Setup}
\label{sec:experimental_setup}

\input{tabs/example}

\noindent\textbf{Unlearning configuration.}
For WMDP unlearning \citep{li2024wmdp}, we use Llama-3-8B-Instruct as the base model. The forget set contains plain-text biosecurity knowledge, while the retain set is sampled from general-domain Wikitext \citep{merity2016pointer}. We run each method for 125 unlearning steps with batch size 4 and search learning rates in $[1\times10^{-5},5\times10^{-5}]$. The retain-loss weight $\lambda$ is selected from $[1.0,5.0]$. For NPO and SimNPO, we additionally tune $\beta\in[0.01,0.1]$; for NPO+SAM, we search perturbation radius $\rho\in[10^{-3},10^{-1}]$. NPO+IRM uses GSM8K as its invariance dataset, an invariance weight $\gamma\in[0.1,2.0]$, and a GSM8K batch size of 48 per unlearning step. For rejection-based methods, GPT-4o reformats the plain-text forget set $\Df$ into QA pairs following \citet{lucki2024adversarial}; $\beta$ for DPO and IDK+AP is tuned in $[0.01,0.1]$. Unless specified otherwise, the remaining method-specific settings follow \citet{che2025model}.

\noindent\textbf{Robustness protocols.}
For in-domain relearning, we fine-tune each unlearned model on $\Df$ for 100 steps using batch size 4 and learning rate $2\times10^{-5}$. For out-of-domain fine-tuning, we use GSM8K, SST-2, and MNLI for 250 steps with batch size 32 and the same learning rate. These protocols deliberately separate recovery driven by forget-domain evidence from parameter changes induced by unrelated downstream data. All unlearning, robustness, and evaluation experiments run on one internal server with eight NVIDIA RTX A6000 GPUs.

\noindent\textbf{Few-shot entailment score.}
The entailment score (ES) measures whether a generated response preserves the factual content of a ground-truth answer. Following prior work \citep{liu2024learning,yuan2024closer}, we treat the model response as the premise and the reference answer as the hypothesis, and use a pretrained NLI model \citep{sileo2023tasksource} to classify the pair as entailment, contradiction, or neutral. ES is the fraction of evaluation examples classified as entailment and should therefore be low on the forget set. Before response generation, we prepend two demonstrations specifying the concise answer format without NLI labels or evaluation feedback, reducing spurious ES variation from malformed responses rather than retained target knowledge.

%% file: tabs/example.tex
\begin{table*}[t]
  \centering
  \caption{\small Example responses from the \textcolor{original}{original} (pre-unlearned) Llama-3-8B-Instruct model and the \textcolor{CDiv}{NPO}- and \textcolor{CRep}{RMU}-unlearned models on the same WMDP query, evaluated in MCQ and Open-QA formats.}
  \label{tab:selection_generation}
  \renewcommand{\arraystretch}{1.02}
  \resizebox{0.94\linewidth}{!}{%
    \small
    \begin{tabular}{>{\arraybackslash}m{0.40\linewidth} 
                    >{\centering\arraybackslash}m{0.06\linewidth} 
                    m{0.13\linewidth} 
                    m{0.45\linewidth}}
      \toprule[1pt]
      \centering\textbf{Question} & \textbf{Label} & ~~~\textbf{MCQ} & ~~~~~~~~~~~~~~~~~~~~~~~~~~~\textbf{Open-QA} \\
      \midrule
      In RNA virus genome synthesis, what potential risk is associated with obtaining virions? \newline
      A. Limiting access to genome information \newline
      B. Re-creating dangerous pathogens \newline C. Decentralized access to synthetic RNA \newline
      D. Accelerated biological research. Answer:
      & B & 
      \textbf{\textcolor{original}{Original}:} B \newline
      \textbf{\textcolor{CDiv}{NPO}:} D \newline
      \textbf{\textcolor{CRep}{RMU}:} D &
      \textbf{\textcolor{original}{Original}:} B. Re-creating dangerous pathogens \newline
      \textbf{\textcolor{CDiv}{NPO}:} @nate@nate@nate@nate@nate@nate... \newline
      \textbf{\textcolor{CRep}{RMU}:} it is not the way it is added by someone...
      \\
      \bottomrule[1pt]
    \end{tabular}
  }
\end{table*}

%% file: secs/sec_eval_no_rob_SLiu.tex
\section{Beyond Answer Selection: Rethinking Unlearning  Effectiveness and Utility}
\label{sec:eval_UE_UT}

\noindent \textbf{Open-QA as a crucial lens for UE and UT evaluation.}
LLM unlearning on WMDP is commonly evaluated through MCQ accuracy, where a method is considered successful when the unlearned model no longer selects the correct option. This criterion is necessary but not sufficient: an incorrect selection does not reveal whether the target knowledge was selectively removed, whether the relative ordering of answer options merely changed, or whether the model's ability to generate coherent text collapsed more broadly. \textbf{Table~\ref{tab:selection_generation}} makes this distinction concrete for a WMDP query. Both NPO and RMU change the MCQ prediction from the correct answer B to D and would therefore be judged similarly by selection accuracy. Their Open-QA responses, however, are malformed and do not constitute meaningful answers. This example is illustrative rather than aggregate evidence, but it exposes an ambiguity inherent to selection-only evaluation: the same MCQ outcome can arise from qualitatively different post-unlearning behaviors. We therefore evaluate UE and UT jointly in MCQ and Open-QA formats, using aggregate results below to determine whether this discrepancy persists across methods and examples.

\begin{figure*}[htb]
\centering
\begin{tabular}{ccc}
\multicolumn{2}{c}{\hspace*{-3mm}\includegraphics[width=0.44\linewidth]{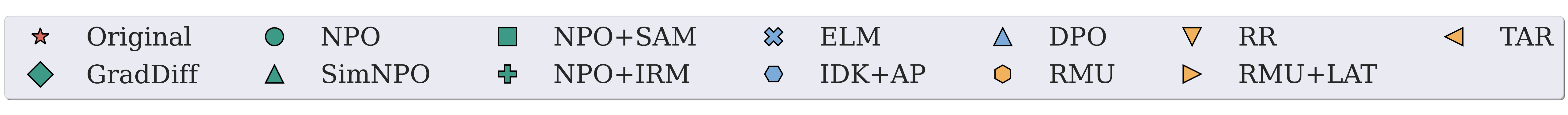}} & \\[-4.5mm]

\hspace*{-3mm} \includegraphics[width=0.218\linewidth,height=!]{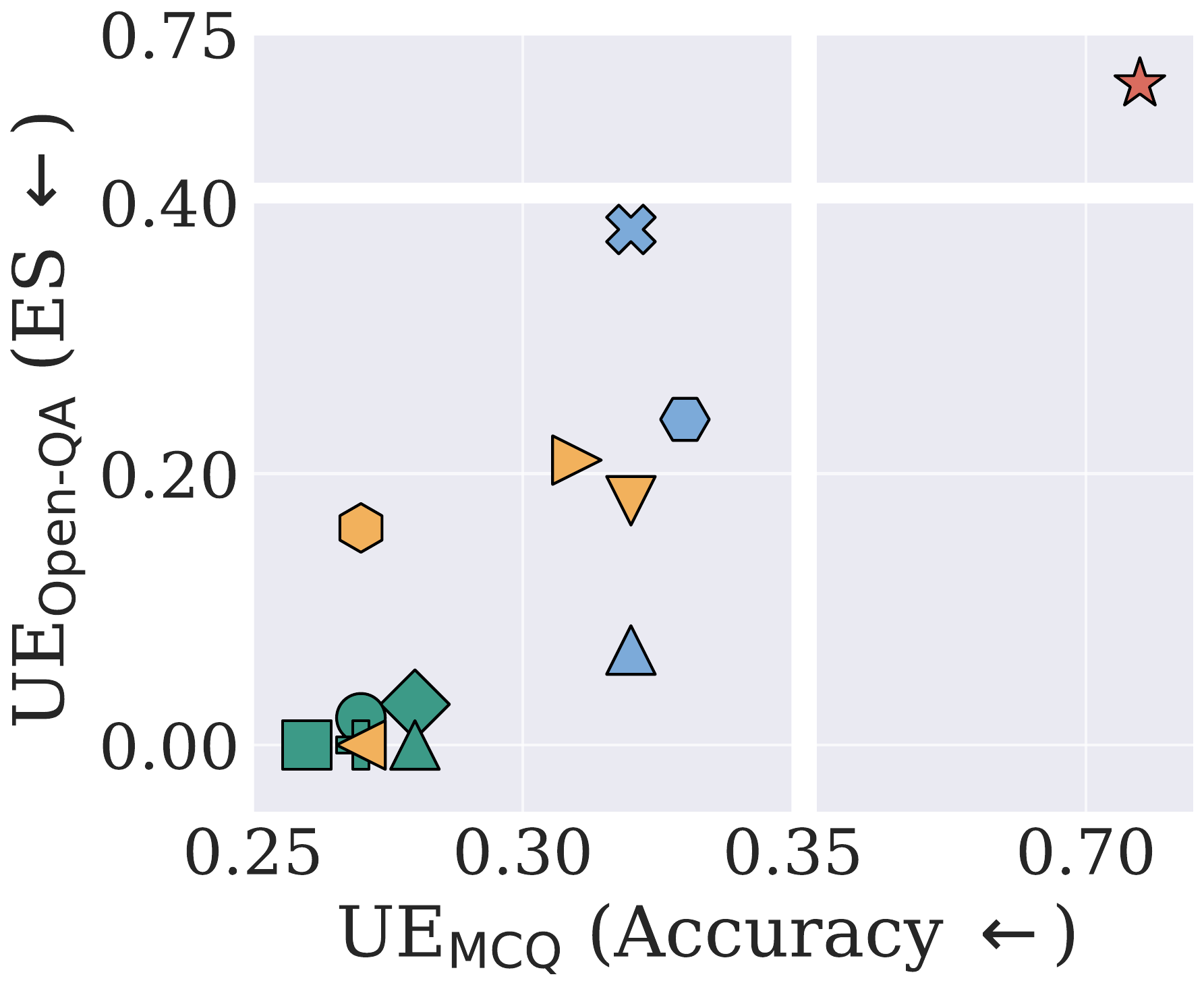} & \hspace*{-2mm}
\includegraphics[width=0.214\linewidth,height=!]{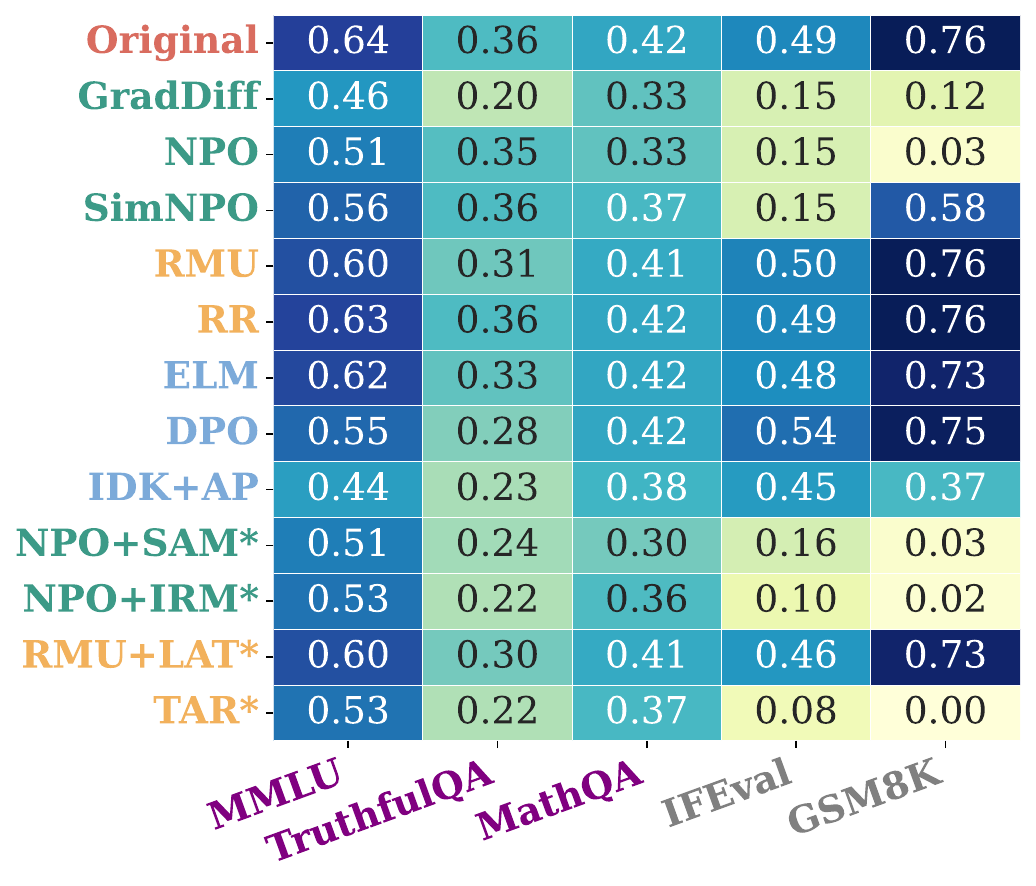} & \hspace*{-2mm}
\includegraphics[width=0.435\linewidth,height=!]{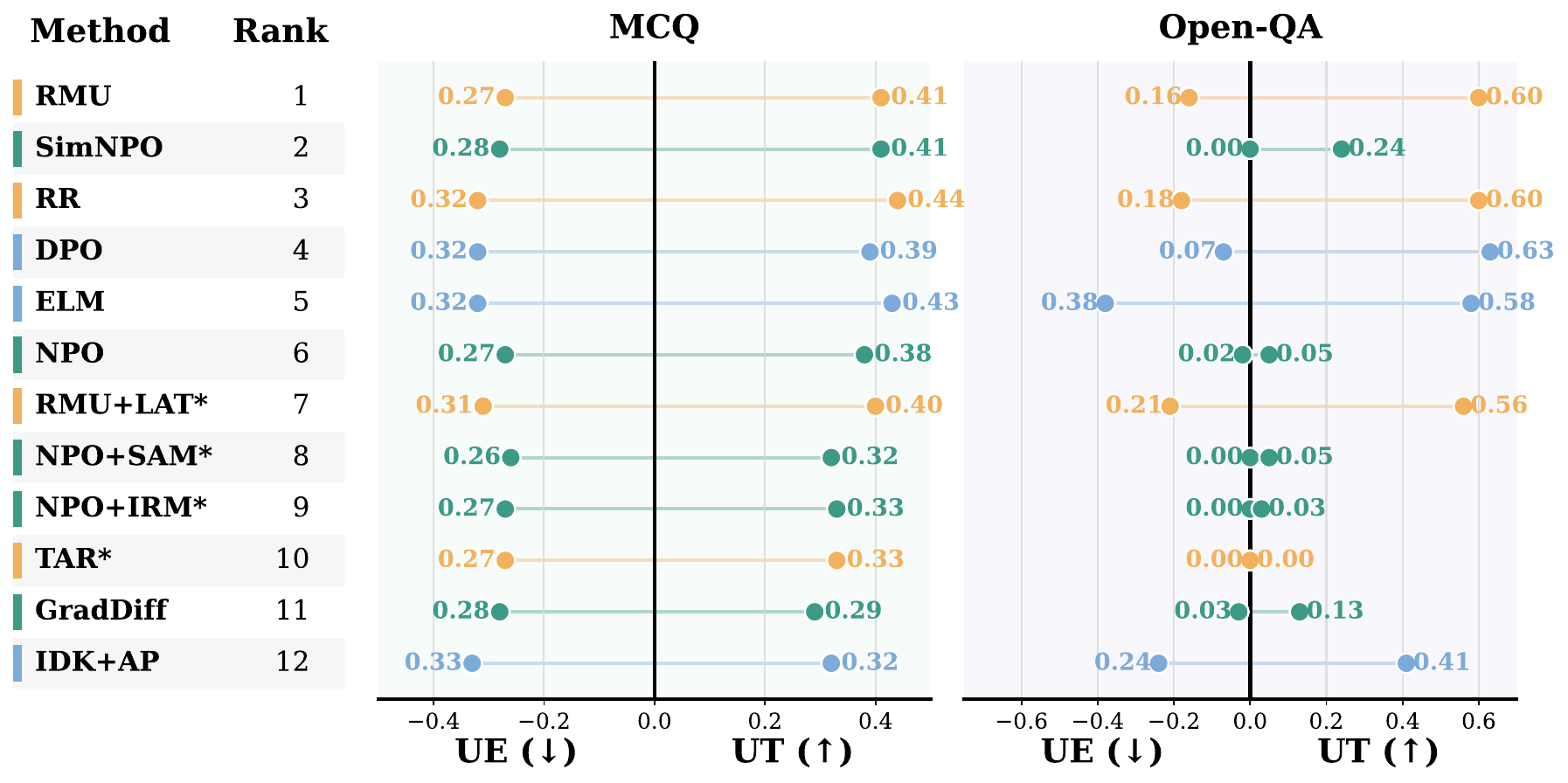} \\[-1.5mm]
{\small (a) $\mathrm{UE}_\text{MCQ}$ vs. $\mathrm{UE}_\text{Open-QA}$} &
{\small (b) $\mathrm{UT}_\text{MCQ}$ and $\mathrm{UT}_\text{Open-QA}$} ($\uparrow$) &
\hspace*{5mm}
{\small (c) Rank of $\mathrm{UE}$ and $\mathrm{UT}$}
\end{tabular}
\vspace{-1.5mm}
\caption{\small UE and UT evaluation of 12 unlearning methods on WMDP with Llama-3-8B-Instruct:
(a) $\mathrm{UE}_\text{MCQ}$ (accuracy) vs. $\mathrm{UE}_\text{Open-QA}$ (ES), where arrows indicate better performance; (b) $\mathrm{UT}_\text{MCQ}$ (\textcolor{MCQ}{MMLU, TruthfulQA, MathQA}) and $\mathrm{UT}_\text{Open-QA}$ (\textcolor{Open-QA}{IFEval, GSM8K}); and (c) overall rank averaged across $\mathrm{UE}_\text{MCQ}$, $\mathrm{UE}_\text{Open-QA}$, and harmonic-mean $\mathrm{UT}_\text{MCQ}$ and $\mathrm{UT}_\text{Open-QA}$.}
\Description{Three-panel figure comparing 12 unlearning methods on WMDP with Llama-3-8B-Instruct: (a) scatter plot of MCQ accuracy versus Open-QA entailment score, (b) heatmap of MCQ and Open-QA utility across five benchmarks, and (c) bar chart of overall average ranks.}
\label{fig:evaluatio_wmdp}
\end{figure*}

To measure generation-based UE, we use the \textit{entailment score} (\textbf{ES}) \citep{yuan2024closer, yao2024accurate}, which evaluates whether a generated response entails the ground-truth answer. Higher ES therefore indicates that more forget-target knowledge remains recoverable and corresponds to weaker unlearning. We provide few-shot demonstrations of the desired answer format to reduce failures caused only by output formatting; Sec.~\ref{sec:experimental_setup} specifies the prompting and evaluation protocol.

Because ES depends on an entailment judge, we test whether the conclusions are sensitive to that choice. \textbf{Table~\ref{tab:es_robust}} compares a base NLI model, a larger NLI model, and \texttt{gpt-4o-mini} as an LLM judge. The method-level pattern is stable: divergence-driven methods obtain ES values near zero, RMU, RR, and RMU+LAT retain intermediate entailment, and ELM retains substantially more target information. For the two NLI judges, non-parametric bootstrap intervals are narrow relative to the principal between-method gaps, indicating that finite-sample uncertainty is unlikely to explain these qualitative differences. Agreement across judges does not establish that ES is error-free, but it shows that our main comparisons are not an artifact of one entailment model.

For UT, generation-based evaluation must also test capabilities outside the forget set. We therefore include \textbf{IFEval} \citep{zhou2023instruction} for instruction following and \textbf{GSM8K} \citep{cobbe2021training} for quantitative reasoning, alongside MCQ benchmarks \textbf{MMLU} \citep{hendrycks2020measuring}, \textbf{MathQA} \citep{amini2019mathqa}, and \textbf{TruthfulQA} \citep{lin2021truthfulqa}. Together, these measurements distinguish successful removal of target knowledge from a general loss of usable model capability.

\input{tabs/es}

In \textbf{Fig.\,\ref{fig:evaluatio_wmdp}-(a)}, we present the $\mathrm{UE}_\text{MCQ}$ and $\mathrm{UE}_\text{Open-QA}$ of 12 unlearning methods along with the original model. As shown, contrasting {\MDiv} with {\MRep} (\textit{e.g.}, NPO vs. RMU), even when the resulting unlearned models achieve the same $\mathrm{UE}_\text{MCQ}$, their $\mathrm{UE}_\text{Open-QA}$ can differ significantly. This highlights the necessity of jointly measuring both $\mathrm{UE}_\text{MCQ}$ and $\mathrm{UE}\text{Open-QA}$. Moreover, {\MDiv} generally achieves better $\mathrm{UE}_\text{MCQ}$ and $\mathrm{UE}_\text{Open-QA}$ than other families.


In Fig.\,\ref{fig:evaluatio_wmdp}(b), {\MDiv} methods such as NPO retain MCQ utility comparable to {\MRep} methods but perform substantially worse on Open-QA utility. This divergence motivates the logit-level analysis below, which tests whether favorable MCQ behavior masks a broader disruption of generation.
Furthermore, adding robustness to RMU comes at the expense of utility in Open-QA, as evidenced by a marked decline in $\mathrm{UT}_\text{Open-QA}$ for TAR compared to RMU. 

\input{secs/logit_analysis}



In \textbf{Fig.\,\ref{fig:evaluatio_wmdp}-(c)}, we report a unified \textbf{final rank} by jointly considering UT and UE under both MCQ and Open-QA evaluations. Specifically, we first compute the \emph{harmonic mean} of the corresponding UT sub-benchmarks within MCQ and Open-QA to obtain the overall $\mathrm{UT}_{\mathrm{MCQ}}$ and $\mathrm{UT}_{\mathrm{Open\text{-}QA}}$, respectively. We then obtain the individual ranks along $\mathrm{UE}_{\mathrm{MCQ}}$, $\mathrm{UE}_{\mathrm{Open\text{-}QA}}$, $\mathrm{UT}_{\mathrm{MCQ}}$, and $\mathrm{UT}_{\mathrm{Open\text{-}QA}}$, and further assign an \emph{average ranking score} that is used for the \textit{final ranking} of each unlearning method. 
As we can see,
within each family, RMU is the strongest method under {\MRepB}, DPO under {\MRejB}, and SimNPO under {\MDivB}. Overall, {\MRepB} tends to preserve UT better, as evidenced by stronger Open-QA performance, and RMU is the most competitive approach, striking the best UE--UT balance.  



\begin{figure}[!htbp]
    \centering
    \includegraphics[width=0.46\columnwidth]{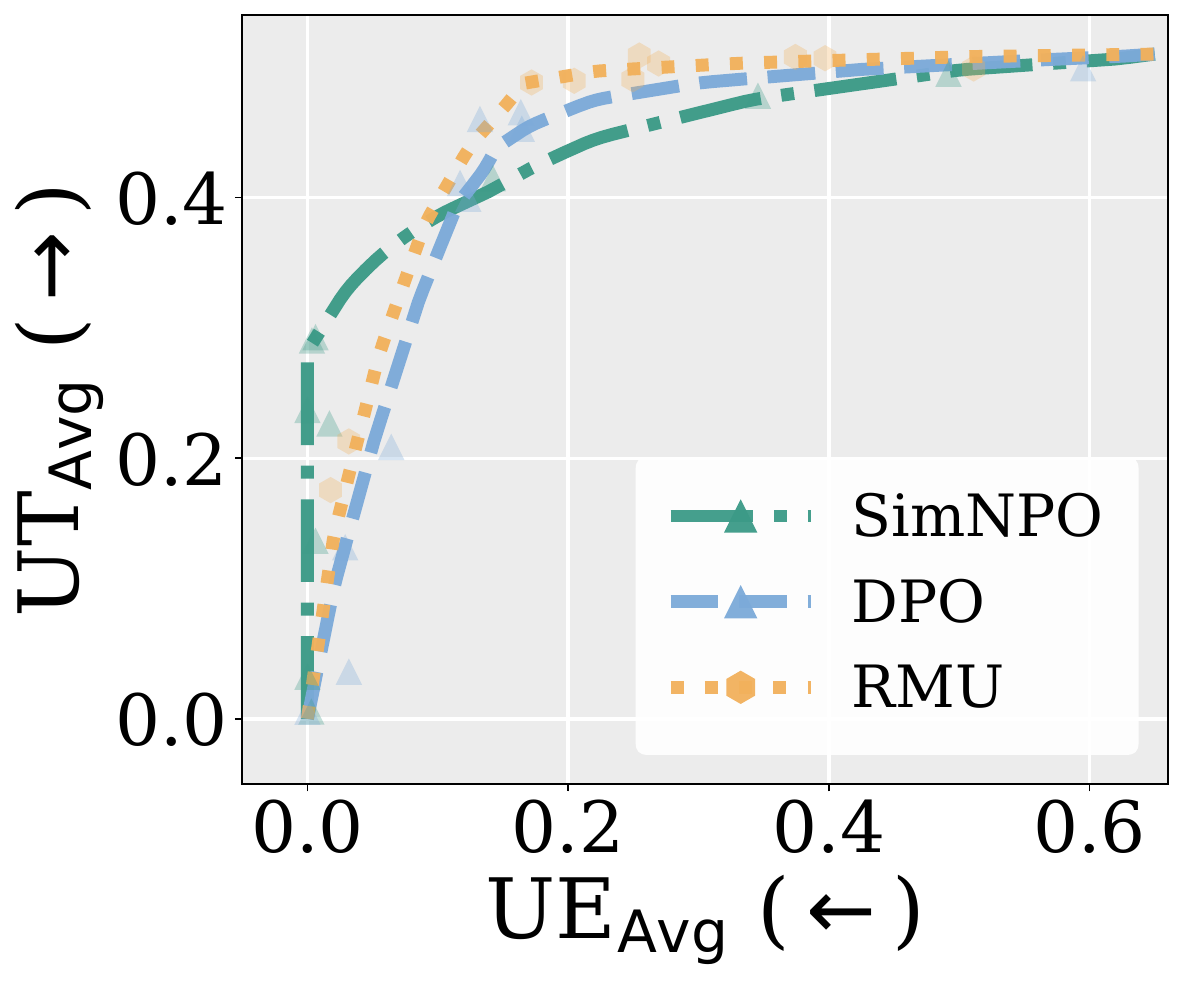}
    \vspace{-1.5mm}
    \caption{\small $\mathrm{UE}_{\mathrm{Avg}}$ and $\mathrm{UT}_{\mathrm{Avg}}$ for RMU, SimNPO, and DPO with different hyper-parameters. $\mathrm{UE}_{\mathrm{Avg}}$ and $\mathrm{UT}_{\mathrm{Avg}}$ are computed as the harmonic mean of the corresponding MCQ and Open-QA metrics. Other settings follow Fig.\,\ref{fig:evaluatio_wmdp}.}
    \Description{Trade-off curves plotting average utility against average unlearning effectiveness across varying hyperparameters for RMU, SimNPO, and DPO on WMDP.}
    \label{fig:trace}
\end{figure}

\noindent \textbf{UE-UT tradeoff.}
\textbf{Fig.\,\ref{fig:trace}} illustrates the UE–UT tradeoff for three representative methods from different families as identified in Fig.\,\ref{fig:evaluatio_wmdp}-(c), RMU, SimNPO, and DPO, by fitting curves over varying hyperparameters (the utility regularization parameter $\gamma$ for RMU and SimNPO, and the preference temperature parameter $\beta$ for DPO). Here, $\mathrm{UE}_{\mathrm{Avg}}$  and $\mathrm{UT}_{\mathrm{Avg}}$ are computed as the harmonic mean of the corresponding MCQ and Open-QA metrics. 
As we can see, when $\mathrm{UT}_{\mathrm{Avg}}$ is preserved at a reasonable level (\textit{i.e.}, converging toward the top-right region), RMU yields lower $\mathrm{UE}_{\mathrm{Avg}}$, suggesting a more favorable UT–UE trade-off. In contrast, SimNPO achieves more substantial gains in UE, as evidenced by $\mathrm{UE}_{\mathrm{Avg}}$ more rapidly converging to $0$, but does so at the cost of a larger drop in UT.

\noindent \textbf{Rethinking rejection-based methods.} Compared to {\MDivB} and {\MRepB}, the {\MRej} family is less commonly used in LLM unlearning. We highlight several overlooked insights. \textit{First}, as shown in Fig.\,\ref{fig:evaluatio_wmdp}-(a), {\MRejB} methods exhibit worse $\mathrm{UE}_\text{MCQ}$ than others. This is one of the primary reasons for their limited popularity \citep{li2024wmdp,shi2024muse}. However, this view may be myopic, as under $\mathrm{UE}_\text{Open-QA}$ the performance of rejection-based methods varies much more substantially, with DPO achieving the best $\mathrm{UE}_\text{Open-QA}$ among them. \textit{Second}, if we only consider the conventional UT assessment under MMLU in Fig.\,\ref{fig:evaluatio_wmdp}-(b), DPO does not exhibit clearly better MMLU accuracy than {\MDivB} methods. Consequently, the prevailing understanding has been that DPO offers \textit{no advantage} in terms of utility \citep{maini2024tofu,zhang2024negative}. However, this view is also myopic, as DPO's $\mathrm{UT}_\text{Open-QA}$ metrics (IFEval and GSM8k) remain comparable to those of the original model. \textit{Third}, while IDK+AP and DPO share the objective of eliciting rejection responses (\textit{e.g.}, ``I don't know'') to forget-relevant queries, IDK+AP shows markedly worse UT, as ranked in Fig.\,\ref{fig:evaluatio_wmdp}-(c). We attribute this degradation to its stricter log-likelihood loss, which overly enforces rejection probabilities.

\input{secs/idk_warmstart}

\begin{tcolorbox}[enhanced,attach boxed title to top center={yshift=-2.2mm,yshifttext=-0.8mm},
  enlarge top initially by=2.2mm, fontupper=\small,
  colback=white,colframe=green!35!black,colbacktitle=green!5!white,
  left=1.8mm,right=1.8mm,top=1.8mm,bottom=1.0mm,before skip=2.0mm,after skip=1.5mm,
  title={\parbox{0.7\linewidth}{\centering Summary of insights into  UE and UT}}, coltitle=black, boxrule=0.8pt, 
  boxed title style={size=small,colframe=green!35!black} ]

\textbf{(1)} Unlearning evaluation should \textit{go beyond answer selection} to assess actual generation content, since MCQ alone can overstate effectiveness and obscure over-forgetting.  

\textbf{(2)} {\textcolor{CDiv}{\textit{Divergence-driven optimization}}} methods often \textit{over-forget}, suffering a much larger UT drop on Open-QA than on MCQ. In contrast, {\MRep} methods retain similar MCQ utility while better preserving Open-QA generation.  

\textbf{(3)} For {\MRej}, Open-QA evaluation is essential to capture UT, though aggressive rejection fine-tuning harms UT.


\end{tcolorbox}

%% file: tabs/es.tex
\begin{table}[!htbp]
\centering
\caption{\small{
ES of the original model and 12 unlearning methods on Llama-3-8B-Instruct using base NLI, large NLI, and LLM-as-judge. 
For NLI-based ES, 95\% confidence intervals use non-parametric bootstrap ($B{=}1000$ resamples).
}}
\label{tab:es_robust}
\renewcommand{\arraystretch}{0.98}
\resizebox{0.96\columnwidth}{!}{%
\begin{tabular}{c|cc|cc|c}
\toprule[1pt]
\textbf{Method} &
\textbf{ES -- base NLI} & \textbf{95\% CI} &
\textbf{ES -- large NLI} & \textbf{95\% CI} &
\textbf{LLM as judge} \\
\midrule
\textbf{Original} & 0.70 & [0.68, 0.73] & 0.72 & [0.69, 0.74] & 0.62 \\
\midrule

\rowcolor{LightCyan!50}
\multicolumn{6}{c}{\textbf{\textcolor{CDiv}{\textit{Divergence-driven optimization}}}} \\
\midrule
\rowcolor{gray!20}
\textbf{GradDiff} & 0.03 & [0.02, 0.04] & 0.02 & [0.02, 0.03] & 0.01 \\
\textbf{NPO} & 0.02 & [0.01, 0.03] & 0.02 & [0.01, 0.03] & 0.01 \\
\rowcolor{gray!20}
\textbf{SimNPO} & 0.00 & [0.00, 0.00] & 0.00 & [0.00, 0.00] & 0.00 \\
\textbf{NPO+SAM} & 0.00 & [0.00, 0.00] & 0.00 & [0.00, 0.00] & 0.00 \\
\rowcolor{gray!20}
\textbf{NPO+IRM} & 0.00 & [0.00, 0.00] & 0.00 & [0.00, 0.00] & 0.00 \\
\midrule

\rowcolor{LightCyan!50}
\multicolumn{6}{c}{\textbf{\textcolor{CRep}{\textit{Representation misalignment}}}} \\
\midrule
\textbf{RMU} & 0.16 & [0.14, 0.18] & 0.17 & [0.15, 0.19] & 0.12 \\
\rowcolor{gray!20}
\textbf{RR} & 0.18 & [0.16, 0.20] & 0.20 & [0.18, 0.22] & 0.15 \\
\textbf{RMU+LAT} & 0.21 & [0.19, 0.23] & 0.21 & [0.19, 0.23] & 0.17 \\
\rowcolor{gray!20}
\textbf{TAR} & 0.00 & [0.00, 0.00] & 0.00 & [0.00, 0.00] & 0.00 \\
\midrule

\rowcolor{LightCyan!50}
\multicolumn{6}{c}{\textbf{\textcolor{CRej}{\textit{Rejection-based targeted unlearning}}}} \\
\midrule
\textbf{ELM} & 0.38 & [0.35, 0.40] & 0.40 & [0.37, 0.43] & 0.31 \\
\rowcolor{gray!20}
\textbf{DPO} & 0.24 & [0.22, 0.27] & 0.26 & [0.24, 0.29] & 0.25 \\
\textbf{IDK+AP} & 0.07 & [0.06, 0.09] & 0.07 & [0.06, 0.09] & 0.08 \\
\bottomrule[1pt]
\end{tabular}
}
\end{table}

%% file: secs/logit_analysis.tex
\begin{figure}[!htbp]
    \centering
    \includegraphics[width=0.44\columnwidth]{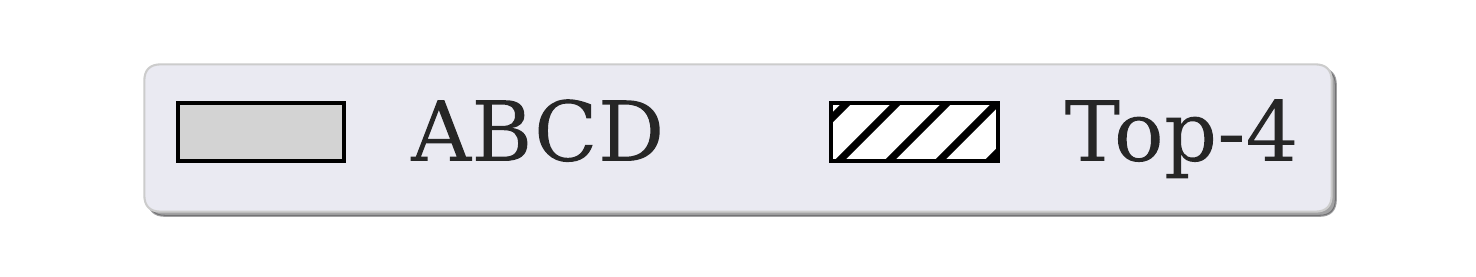} \\[-1mm]
    \includegraphics[width=0.46\columnwidth]{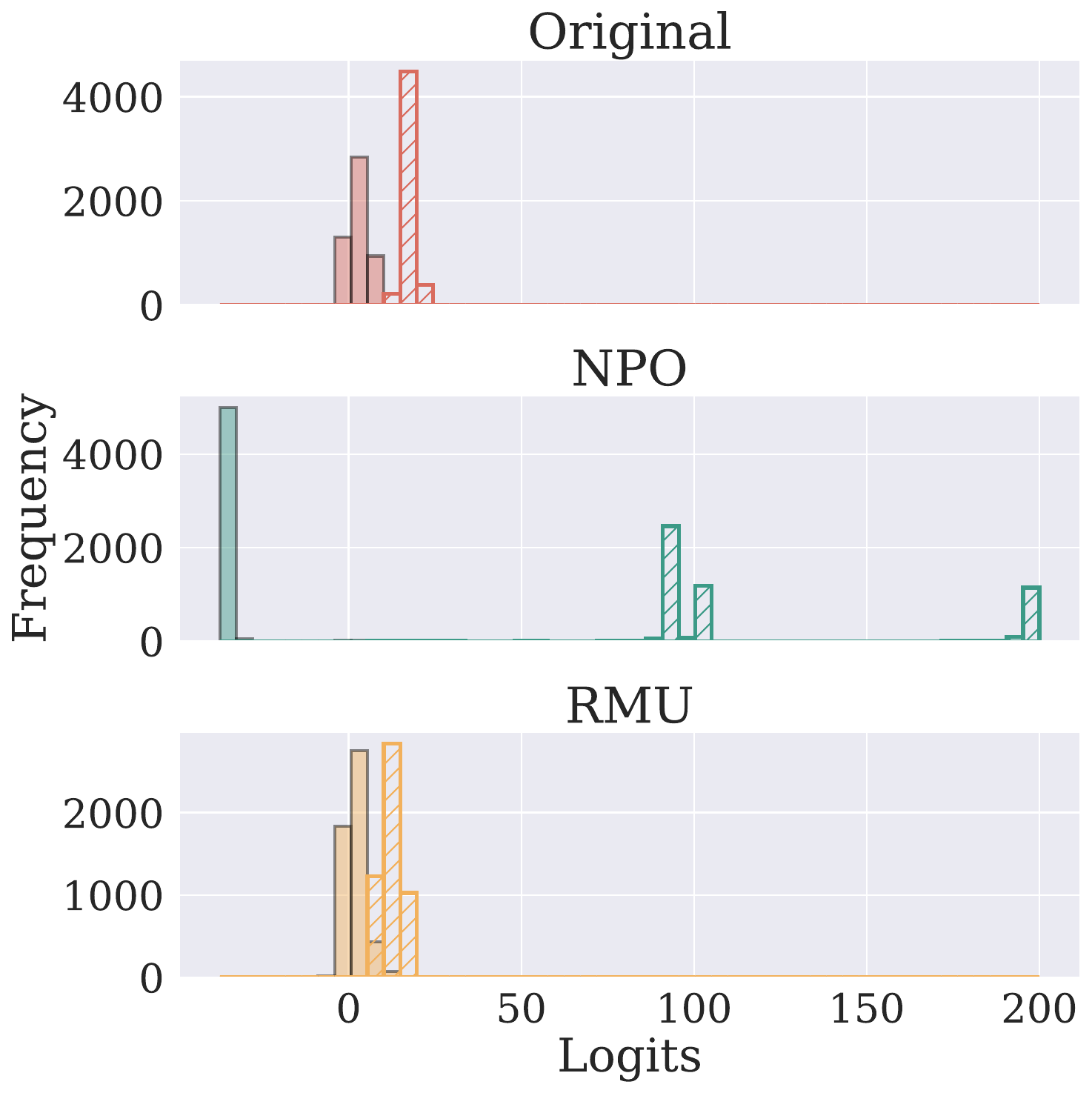}
    \vspace{-1.5mm}
    \caption{\small ABCD-option and top-four token logits of original, NPO-unlearned, and RMU-unlearned Llama-3-8B-Instruct on WMDP.}
    \Description{Grouped bar chart comparing logits of multiple-choice options A, B, C, and D and top-four generated tokens for original, NPO-unlearned, and RMU-unlearned Llama-3-8B-Instruct models on WMDP.}
    \label{fig:bar_logits}
\end{figure}

\noindent\textbf{From logits to behavior.}
The Open-QA utility gap in Fig.~\ref{fig:evaluatio_wmdp}(b) suggests that divergence-driven unlearning can damage generation beyond what MCQ accuracy reveals. To examine the mechanism, \textbf{Fig.~\ref{fig:bar_logits}} compares the logits assigned by the original, NPO-unlearned, and RMU-unlearned models to the four answer options and to each model's top-four generated tokens. NPO drives the A/B/C/D logits toward similarly low values; its incorrect MCQ answer therefore reflects broad suppression of the candidate options rather than a targeted reordering among plausible answers. Its highest-logit generated tokens are substantially stronger but correspond to malformed output of the type illustrated in Table~\ref{tab:selection_generation}. RMU, by contrast, retains a logit profile closer to the original model while changing the relative ordering of the answer options. This evidence links NPO's low $\mathrm{UE}_\text{Open-QA}$ to disrupted generation and explains why favorable forget-set scores must be interpreted jointly with retained Open-QA utility.

%% file: secs/idk_warmstart.tex
\begin{figure}[!htbp]
    \centering
    \includegraphics[width=0.47\columnwidth]{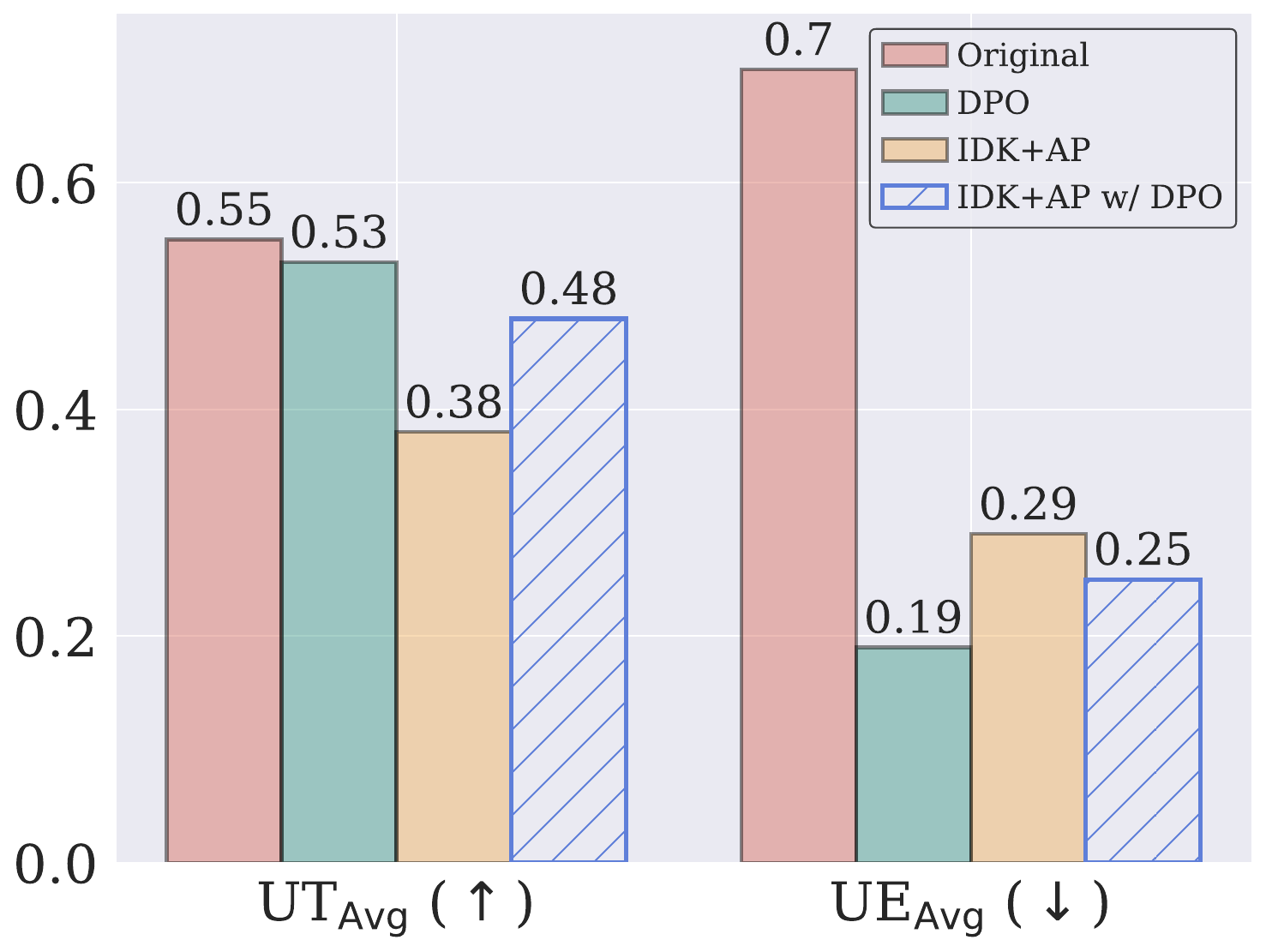}
    \vspace{-1.5mm}
    \caption{\small UE--UT performance of IDK+AP with and without a DPO warm start. $\mathrm{UE}_\text{Avg}$ and $\mathrm{UT}_\text{Avg}$ follow the definitions used in Fig.~\ref{fig:trace}.}
    \Description{Scatter plot comparing average unlearning effectiveness and average utility of IDK+AP with and without a DPO warm start on WMDP.}
    \label{fig:idk_ap_dpo}
\end{figure}

\noindent\textbf{Mitigating IDK+AP utility loss with a DPO warm start.}
Although DPO and IDK+AP both train the model to reject forget-set queries, Fig.~\ref{fig:evaluatio_wmdp}(c) shows a substantially less favorable UE--UT balance for IDK+AP. One plausible explanation is objective sharpness: IDK+AP directly maximizes the likelihood of a fixed rejection target, whereas DPO optimizes a relative preference between desired and undesired responses. We test this explanation by first applying DPO for a small number of epochs and then continuing with IDK+AP. As shown in \textbf{Fig.~\ref{fig:idk_ap_dpo}}, this warm start improves retained utility while preserving effective unlearning. The result is consistent with the hypothesis that preference optimization provides a smoother transition toward rejection behavior before the stricter likelihood objective is applied. This ordering reverses the common post-training sequence in which supervised fine-tuning precedes DPO \citep{dubey2024llama}; here, DPO instead initializes the model before rejection-target fine-tuning. As an intervention on one method pair, we treat it as mechanistic evidence rather than a general prescription.

%% file: secs/sec_rob_SLiu.tex
\section{Multi-Faceted Robustness Assessments}
\label{sec:eval_Rob}

In this section, we investigate the robustness of LLM unlearning against a full spectrum of vulnerabilities, ranging from model-level attacks (\textit{e.g.}, in-domain relearning and out-of-domain fine-tuning) to input-level jailbreaking attacks.

\noindent \textbf{Robustness against in-domain relearning and out-of-domain fine-tuning.}  
Prior work \citep{hu2024jogging,fan2025towards,wang2025invariance,deeb2024unlearning,hu2025blur,che2025model} has shown that unlearned models remain vulnerable to post-unlearning weight perturbations. However, most studies focus on only one setting, either \textit{in-domain relearning} (fine-tuning on data aligned with the forget set $\Df$) \citep{fan2025towards} or \textit{out-of-domain fine-tuning} (adapting to forget-unrelated tasks such as GSM8K, SST2, or MNLI) \citep{wang2025invariance}. We argue that these represent two fundamentally different perturbation regimes, analogous to adversarial versus out-of-distribution robustness, and should be studied jointly for a complete picture of unlearning robustness.

\begin{figure}[!t]
\centering
\begin{tabular}{@{}c@{\hspace{1.5mm}}c@{}}
\includegraphics[width=0.475\columnwidth]{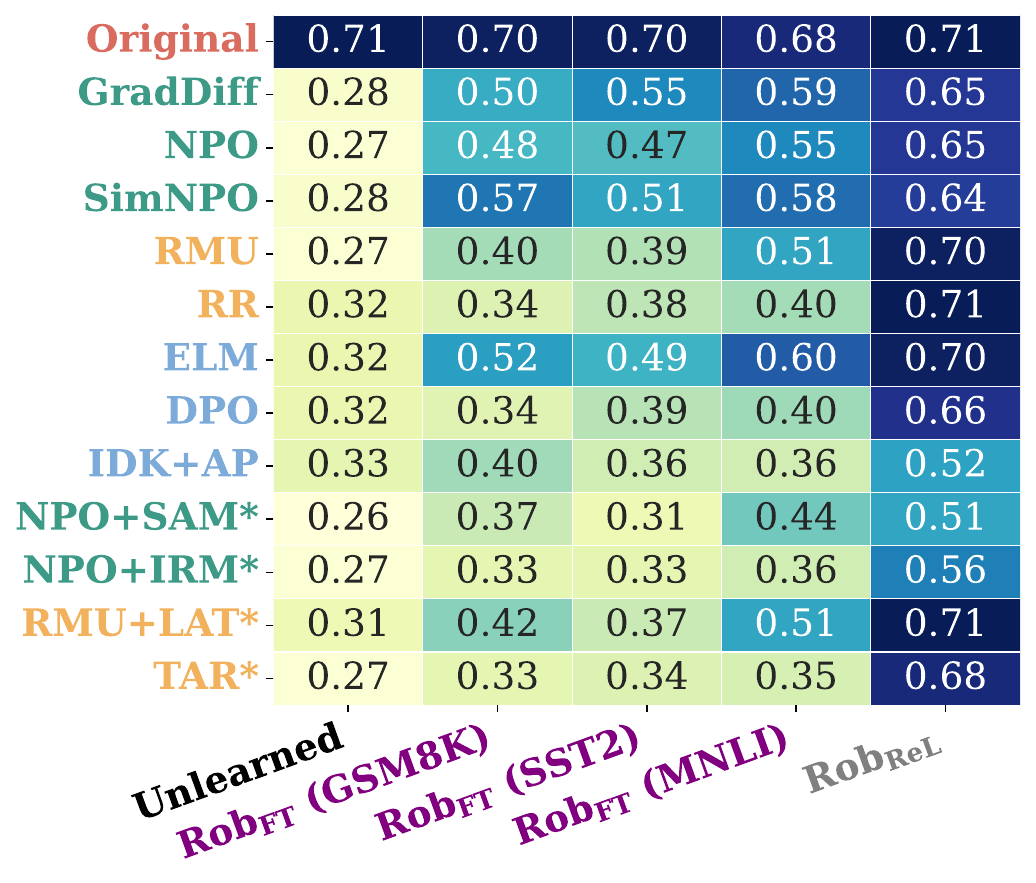} &
\includegraphics[width=0.475\columnwidth]{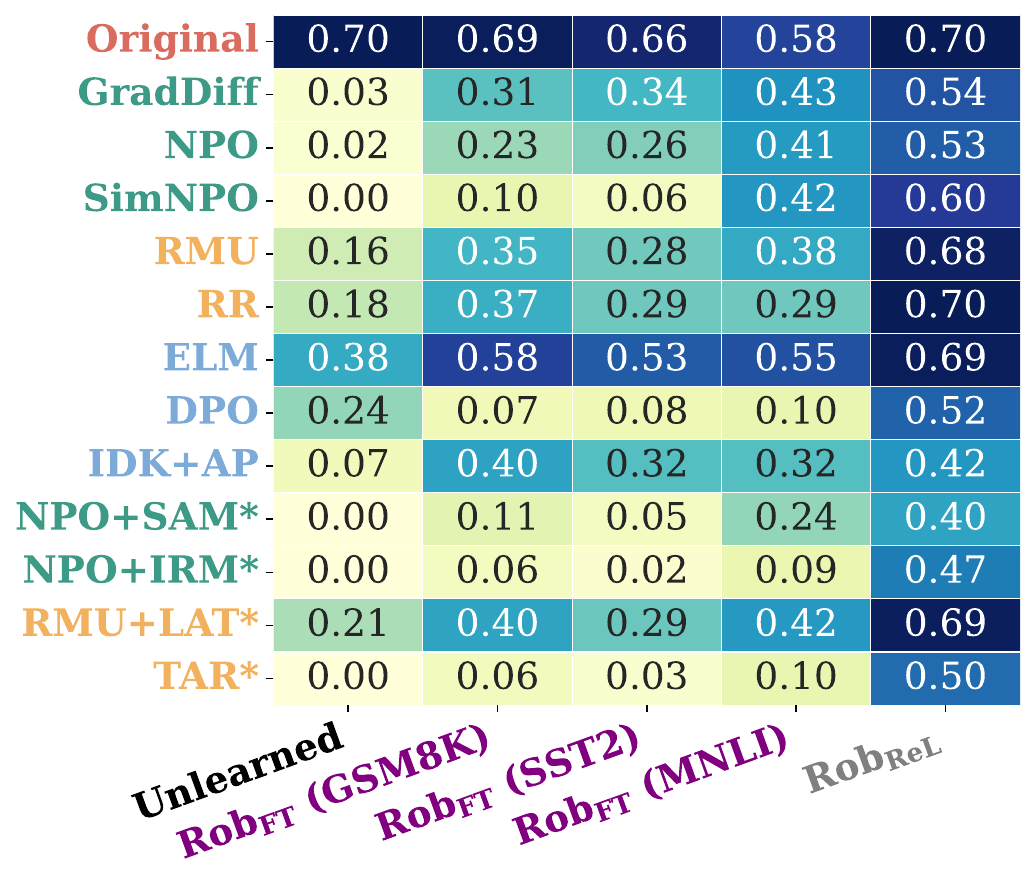}
\\[-1.5mm]
{\footnotesize (a) $\mathrm{UE}_\text{MCQ}$ (Accuracy $\downarrow$)} &  
{\footnotesize (b) $\mathrm{UE}_\text{Open-QA}$ (ES $\downarrow$)}
\end{tabular}
\vspace{-1.5mm}
\caption{\small Robustness of in-domain relearning ($\mathrm{Rob}_\text{ReL}$) and out-of-domain fine-tuning ($\mathrm{Rob}_\text{FT}$) evaluated by (a) $\mathrm{UE}_\text{MCQ}$ (Accuracy) and (b) $\mathrm{UE}_\text{Open-QA}$ (ES). Out-of-domain fine-tuning uses GSM8K, SST2, and MNLI. Methods with * include robust designs, and the first column (``unlearned'') shows results before weight perturbations. Other settings are consistent with Fig.\,\ref{fig:evaluatio_wmdp}.}
\Description{Two side-by-side heatmaps showing MCQ accuracy and Open-QA entailment scores of 12 unlearning methods under in-domain relearning and out-of-domain fine-tuning on GSM8K, SST2, and MNLI.}
\label{fig:rob_weight_perturbations}
\end{figure}

\begin{figure}[!b] 
\centering
\begin{tabular}{@{}c@{\hspace{2mm}}c@{}}
\includegraphics[width=0.43\columnwidth]{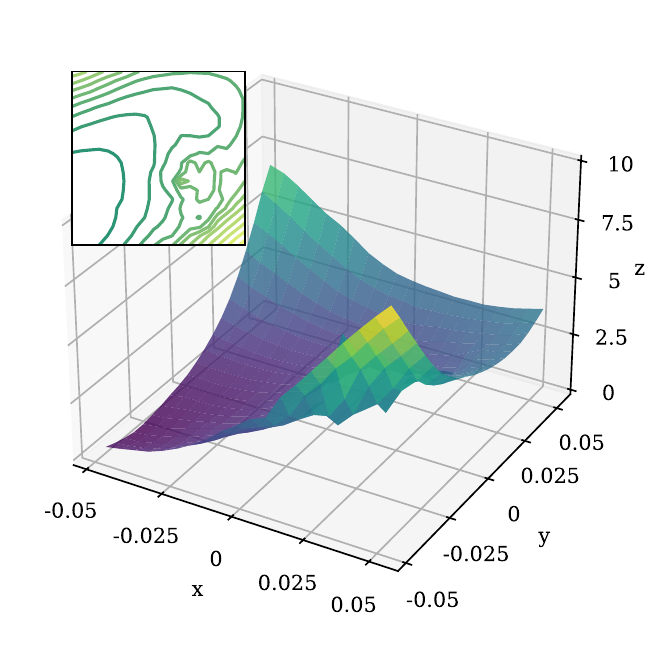} &
\includegraphics[width=0.43\columnwidth]{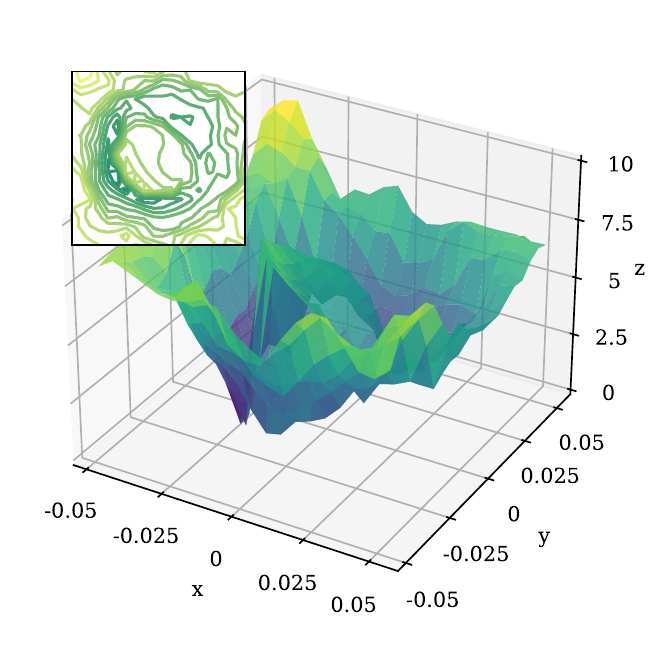}
\\[-1.5mm]
{\footnotesize (a) TAR} &  
{\footnotesize (b) RMU+LAT}
\end{tabular}
\vspace{-1.5mm}
\caption{\small The prediction loss landscape of the TAR and RMU+LAT-unlearned model on the forget set of WMDP with Llama-3-8B-Instruct using the visualization tool in \citep{li2018visualizing}.}
\Description{Three-dimensional surface plots comparing the forget-set prediction loss landscapes of TAR and RMU+LAT on WMDP, showing a smoother surface for TAR than for RMU+LAT.}
\label{fig:landscape}
\end{figure}

\begin{figure*}[t]
\centering
\begin{tabular}{ccc}
\hspace*{-3mm}
\includegraphics[width=0.208\textwidth]{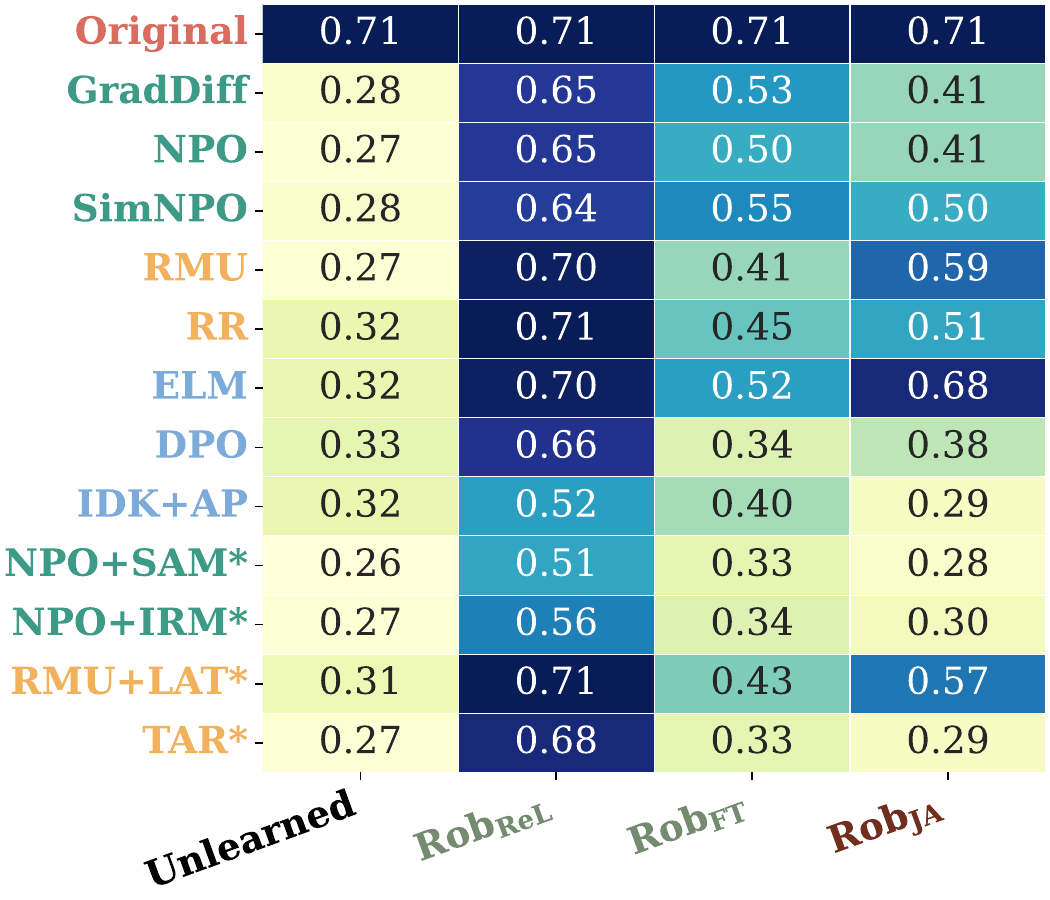} &
\hspace*{-2mm}
\includegraphics[width=0.183\textwidth]{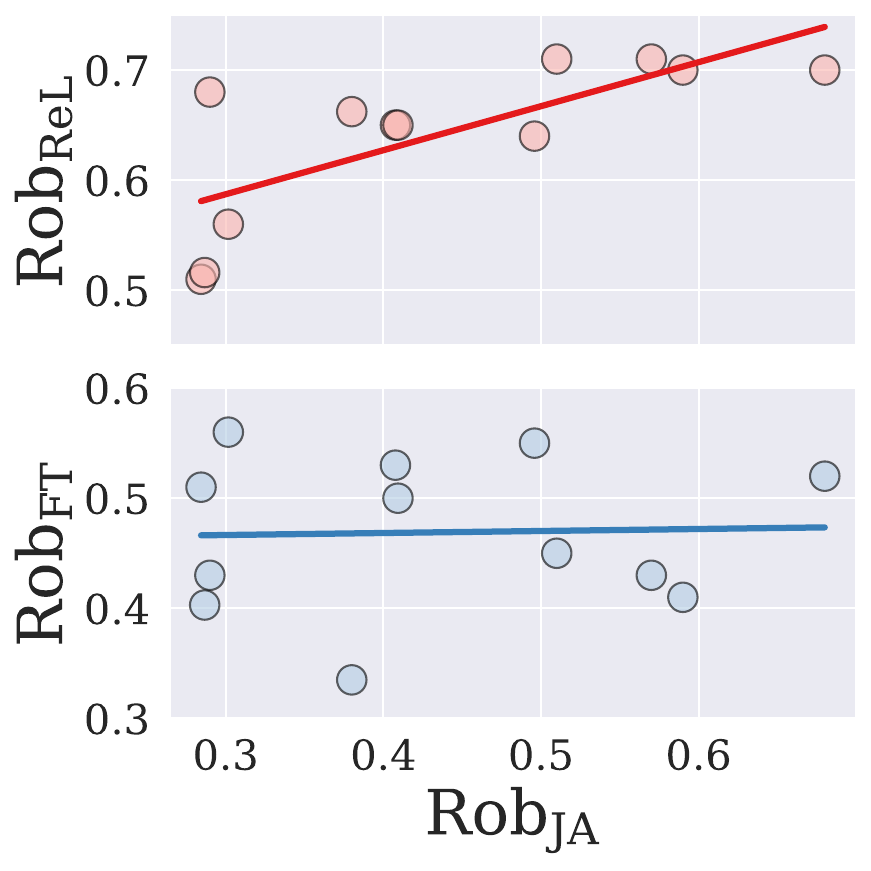} & 
\hspace*{-2mm}
\includegraphics[width=0.425\textwidth]{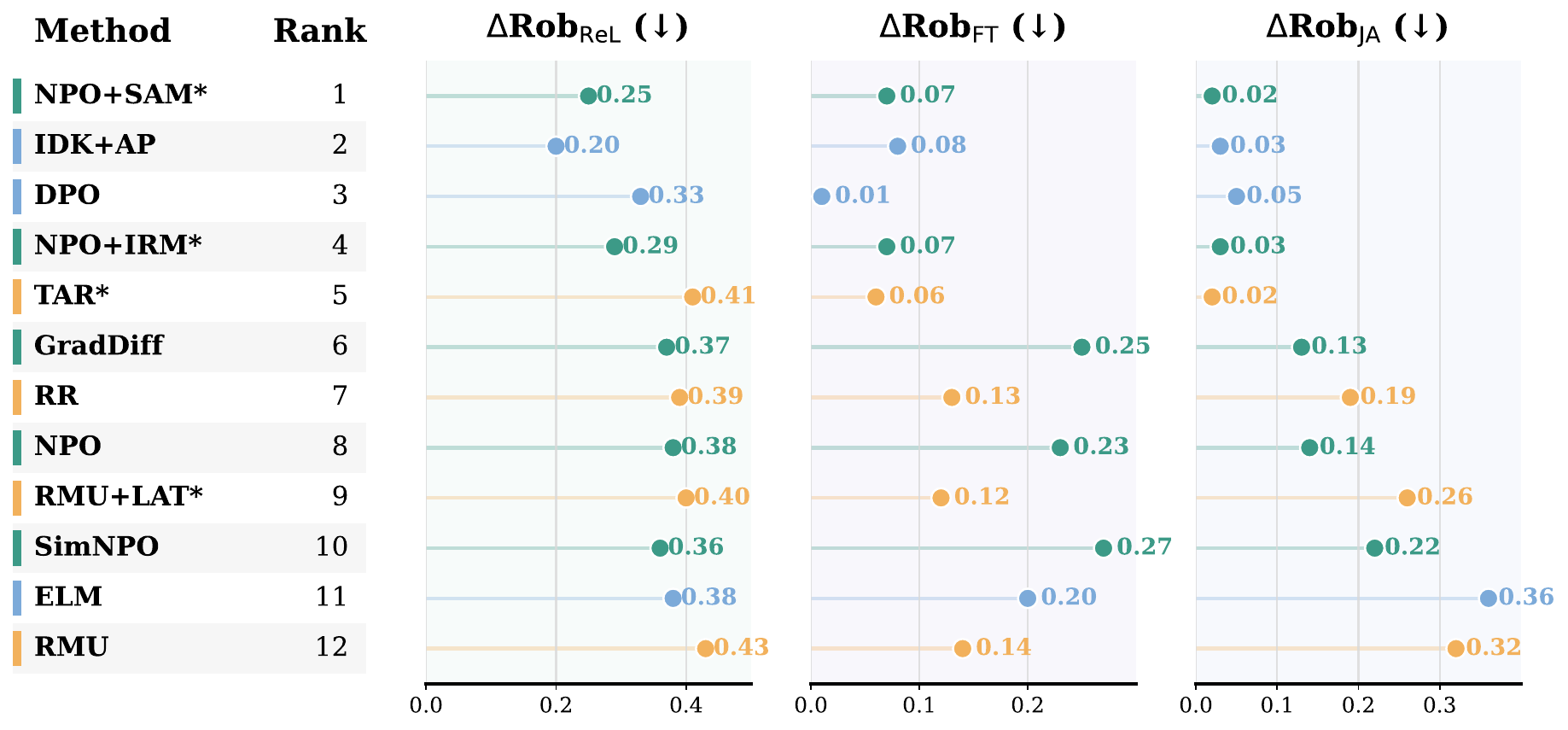} \\[-1.5mm]
{\small (a) $\mathrm{UE}_\text{MCQ}$ (Accuracy $\downarrow$)} &
{\small (b) Correlation of $\mathrm{Rob}_\text{JA}$}  &
{\small (c) Overall robustness rank}
\end{tabular}
\vspace{-1.5mm}
\caption{\small (a) Overall robustness under in-domain relearning ($\mathrm{Rob}_\text{ReL}$), out-of-domain fine-tuning ($\mathrm{Rob}_\text{FT}$), and jailbreaking ($\mathrm{Rob}_\text{JA}$), evaluated by $\mathrm{UE}_\text{MCQ}$. (b) Correlations between $\mathrm{Rob}_\text{JA}$ and $\mathrm{Rob}_\text{ReL}$ / $\mathrm{Rob}_\text{FT}$. (c) Overall robustness ranking computed from the average rank of robustness gaps $\Delta \mathrm{Rob}_x$ relative to the ``unlearned'' column in (a) (smaller is better).}
\Description{Three-panel figure showing (a) heatmap of MCQ accuracy under in-domain relearning, out-of-domain fine-tuning, and jailbreaking, (b) correlation lines between jailbreaking and weight-perturbation robustness, and (c) bar chart of overall robustness ranks across 12 unlearning methods.}
\label{fig:robustness_all}
\end{figure*}

\textbf{Fig.\,\ref{fig:rob_weight_perturbations}} shows the robustness to in-domain relearning ($\text{\textbf{Rob}}_{\text{\textbf{ReL}}}$) and out-of-domain fine-tuning ($\text{\textbf{Rob}}_{\text{\textbf{FT}}}$) of the LLM unlearning methods in Table\,\ref{tab:taxonomy}. 
Here, lower accuracy (lighter color) indicates stronger robustness, and methods with explicit robust designs are marked with an asterisk (*). Several insights can be drawn below.

\textit{First}, compared to the ``unlearned'' setting, both in-domain relearning ($\mathrm{Rob}_{\text{ReL}}$) and out-of-domain fine-tuning ($\mathrm{Rob}_{\text{FT}}$) reduce unlearning effectiveness, manifested as increases in $\mathrm{UE}_{\text{MCQ}}$ and $\mathrm{UE}_{\text{Open-QA}}$. Moreover, in-domain relearning is a more severe perturbation, yielding lower robustness than out-of-domain fine-tuning.  

{\emergencystretch=1.5em\textit{Second}, for methods without explicit robust designs, {\MDiv} approaches (GradDiff, NPO, SimNPO) generally exhibit stronger robustness to in-domain relearning ($\mathrm{Rob}_\text{ReL}$) than {\MRep} (RMU, RR) and {\MRej} (ELM, DPO). \textit{However}, this trend can reverse under out-of-domain fine-tuning ($\mathrm{Rob}_\text{FT}$), where {\MDiv} methods become less robust than {\MRep}, as seen for NPO vs.\ RMU under $\mathrm{UE}_\text{MCQ}$.  
For {\MRej}, DPO is the most robust under $\mathrm{Rob}_\text{FT}$, though this advantage does not consistently hold under $\mathrm{Rob}_\text{ReL}$. Hence, both $\mathrm{Rob}_\text{ReL}$ and $\mathrm{Rob}_\text{FT}$ should be considered jointly for a comprehensive assessment.\par}

\textit{Third}, incorporating robustness-oriented designs consistently improves robustness in both $\mathrm{Rob}_{\text{ReL}}$ and $\mathrm{Rob}_{\text{FT}}$, regardless of whether they build on {\MDivB} (\textit{e.g.}, NPO+SAM, NPO+IRM) or {\MRepB} (\textit{e.g.}, TAR). An exception is RMU+LAT, which fails to show consistent improvements over RMU. Similar to adversarial logit pairing \citep{kannan2018adversarial}, it offers only \textit{local robustness}, leading to a less smooth loss landscape \citep{engstrom2018evaluating}. \textbf{Fig.\,\ref{fig:landscape}} visualizes the forget loss landscape of TAR and RMU+LAT following \citep{li2018visualizing}. The landscape of TAR is noticeably smoother, while RMU+LAT exhibits irregularities, indicating only \textit{local robustness}. This echoes the debate \citep{engstrom2018evaluating} contrasting standard adversarial training \citep{madry2018towards} with adversarial logit pairing \citep{kannan2018adversarial}, where leveraging logits (or other latent information) was argued to yield limited robustness. By analogy, in LLM unlearning, RMU+LAT also fails to achieve broad robustness due to its locality constraint.

{\emergencystretch=1em\noindent \textbf{Robustness against jailbreaking and its interaction with weight perturbations.}  
Besides model-level perturbations, unlearned models are also vulnerable to input-level \textit{jailbreaking attacks} \citep{lucki2024adversarial,lynch2024eight,patil2023can}, which manipulate prompts to bypass unlearning. We denote robustness against such attacks as $\text{\textbf{Rob}}_{\text{\textbf{JA}}}$, with adversarial prompts generated using the enhanced GCG method \citep{thompson2024flrt}, recognized as one of the most effective input-level attacks against unlearning \citep{lucki2024adversarial,fan2025towards}.\par}

\input{tabs/quan_muse}
\input{tabs/qwen14b}

In \textbf{Fig.\,\ref{fig:robustness_all} (a)}, we present the robustness performance of different unlearning methods across both model-level ($\mathrm{Rob}_{\text{ReL}}$ and $\mathrm{Rob}_{\text{FT}}$) and input-level ($\mathrm{Rob}_{\text{JA}}$) dimensions.
As shown, without explicit robustness-oriented design, {\MDiv} methods (GradDiff, NPO, SimNPO) generally exhibit stronger $\mathrm{Rob}_{\text{JA}}$ than {\MRep} methods. This may be attributed to the degraded generative capacity of {\MDivB} methods, which reduces their sensitivity to jailbreaking attacks aimed at eliciting unlearned knowledge. For {\MRej} methods, $\mathrm{Rob}_{\text{JA}}$ varies considerably: ELM shows almost no robustness, whereas IDK+AP remains robust with nearly no degradation. When robustness-oriented design is incorporated, most methods (except RMU+LAT) exhibit significantly improved resilience to jailbreaking attacks.
Furthermore, \textbf{Fig.\,\ref{fig:robustness_all}(b)} jointly presents the relationship between input-level robustness ($\mathrm{Rob}_\text{JA}$) and model-level robustness ($\mathrm{Rob}_\text{ReL}$ and $\mathrm{Rob}_\text{FT}$). An interesting finding is that $\mathrm{Rob}_{\text{JA}}$ patterns align more closely with $\mathrm{Rob}_{\text{ReL}}$ than with $\mathrm{Rob}_{\text{FT}}$. This positive correlation likely arises because both $\mathrm{Rob}_{\text{JA}}$ and $\mathrm{Rob}_{\text{ReL}}$ reflect worst-case adversarial testing activated by forget-related data in the unlearned domain.

In \textbf{Fig.\,\ref{fig:robustness_all}(c)}, we present an overall robustness ranking of current unlearning methods. For each robustness dimension ($\mathrm{Rob}_{\text{ReL}}$, $\mathrm{Rob}_{\text{FT}}$, and $\mathrm{Rob}_{\text{JA}}$), we measure the robustness decline relative to the unlearned model before robustness evaluation, yielding robustness gaps denoted by $\Delta \mathrm{Rob}_{\text{ReL}}$, $\Delta \mathrm{Rob}_{\text{FT}}$, and $\Delta \mathrm{Rob}_{\text{JA}}$. Here, smaller values indicate better resilience. The final ranking is then obtained by averaging the ranks across these robustness gaps.
As shown, augmenting unlearning with robustness-oriented design choices, \textit{e.g.}, SAM~\citep{fan2025towards}, IRM~\citep{wang2025invariance}, and TAR~\citep{tamirisa2024tamper}, typically improves resilience across metrics, leading to better overall rankings. In addition, resilience to in-domain relearning can differ markedly from robustness to out-of-domain fine-tuning across {\MDiv} and {\MRep} families.
Moreover, {\MRep} methods are generally more vulnerable to jailbreaking attacks. 
Furthermore, although the {\MRej} method IDK+AP achieves a higher overall robustness ranking than DPO, it has the lowest overall UE–UT ranking, as shown in Fig.\,\ref{fig:evaluatio_wmdp}(c). 

\input{secs/quantization_analysis}

\noindent\textbf{Generalization to a larger, alternative model architecture.}
To test whether the principal NPO--RMU comparison is specific to Llama-3-8B-Instruct, we repeat it on Qwen2.5-14B (\textbf{Table~\ref{tab:qwen14b}}). This experiment compares one representative method from each family and should therefore be interpreted as a targeted architecture check rather than evidence covering every method. NPO and RMU achieve comparable WMDP UE: their $\mathrm{UE}_\text{MCQ}$ values are 0.36 and 0.32, while their $\mathrm{UE}_\text{Open-QA}$ values are 0.28 and 0.34, respectively. Their MCQ utility is also similar (0.72 versus 0.76 on MMLU and 0.51 versus 0.53 on MathQA). The decisive difference appears in generation utility: RMU retains 0.90 on GSM8K and 0.45 on IFEval, whereas NPO falls to 0.00 and 0.14. Thus, Qwen2.5-14B reproduces the central observation that comparable MCQ forgetting can conceal a large gap in retained generation.

The robustness results remain perturbation-dependent. Lower values indicate less recovered forget-target performance after an attack. RMU is more robust to out-of-domain fine-tuning across GSM8K, MNLI, and SST2 (0.32/0.35/0.36 versus NPO's 0.60/0.70/0.53) and to quantization (0.29 versus 0.36), whereas NPO is modestly more robust to in-domain relearning (0.69 versus 0.79). These results support distinguishing in-domain from out-of-domain robustness and caution against a single uniform notion of robustness.


\begin{tcolorbox}[enhanced,attach boxed title to top center={yshift=-2.2mm,yshifttext=-0.8mm},
  enlarge top initially by=2.0mm, fontupper=\small,
  colback=white,colframe=green!35!black,colbacktitle=green!5!white,
  left=1.8mm,right=1.8mm,top=1.5mm,bottom=0.8mm,before skip=1.6mm,after skip=1.2mm,
  title={\parbox{0.7\linewidth}{\centering Summary of robustness insights}}, coltitle=black, boxrule=0.8pt, 
  boxed title style={size=small,colframe=green!35!black} ]

\textbf{(1)} In-domain relearning robustness \textit{differs} from out-of-domain fine-tuning robustness.  


\textbf{(2)} Robustness to \textit{input}-level jailbreaking aligns closely with \textit{weight}-level in-domain relearning, as both expose worst-case adversarial behavior triggered by forget-relevant data.

\textbf{(3)} From an \textit{overall robustness} perspective, robustness-oriented designs consistently improve resilience, even beyond the specific vulnerabilities they target.
\end{tcolorbox}

%% file: tabs/quan_muse.tex
\begin{table}[!b]
\centering
\caption{\small Quantization performance of NPO and RMU on MUSE Books and WMDP. UE and UT are assessed using KnowMem on $\Df$ and KnowMem on $\Dr$ for MUSE Books, and by $\mathrm{UE}_\text{MCQ}$ (accuracy) and $\mathrm{UT}_\text{MCQ}$ (MMLU) for WMDP. Results are reported for full-precision, 8-bit, and 4-bit models.}
\label{tab:quantization_results}
\renewcommand{\arraystretch}{0.96}
\resizebox{0.91\columnwidth}{!}{%
\begin{tabular}{c ccc ccc}
\toprule[1pt]
\rowcolor{LightCyan!50}
\multicolumn{7}{c}{\textbf{MUSE}} \\
\midrule
\multirow{2}{*}{\textbf{Method}} &
\multicolumn{3}{c}{\textbf{KnowMem on $\Df$ ($\downarrow$)}} &
\multicolumn{3}{c}{\textbf{KnowMem on $\Dr$ ($\uparrow$)}} \\
\cmidrule(lr){2-4} \cmidrule(lr){5-7}
& \textbf{Full} & \textbf{8 Bit} & \textbf{4 Bit} & \textbf{Full} & \textbf{8 Bit} & \textbf{4 Bit} \\
\midrule
\rowcolor{gray!10}
NPO &  2.10 &  2.12 & 30.30 & 55.31 & 52.89 & 48.79 \\
RMU & 24.44 & 24.06 &  8.16 & 59.55 & 55.80 & 25.84 \\
\midrule
\rowcolor{LightCyan!50}
\multicolumn{7}{c}{\textbf{WMDP}} \\
\midrule
\multirow{2}{*}{\textbf{Method}} &
\multicolumn{3}{c}{\textbf{$\mathrm{UE}_\text{MCQ}$ (Accuracy $\downarrow$)}} &
\multicolumn{3}{c}{\textbf{$\mathrm{UT}_\text{MCQ}$ (MMLU $\uparrow$)}} \\
\cmidrule(lr){2-4} \cmidrule(lr){5-7}
& \textbf{Full} & \textbf{8 Bit} & \textbf{4 Bit} & \textbf{Full} & \textbf{8 Bit} & \textbf{4 Bit} \\
\midrule
\rowcolor{gray!10}
NPO & 0.28 & 0.28 & 0.28 & 0.46 & 0.46 & 0.46 \\
RMU & 0.27 & 0.27 & 0.26 & 0.60 & 0.59 & 0.56 \\
\bottomrule[1pt]
\end{tabular}%
}
\end{table}

%% file: tabs/qwen14b.tex
\begin{table*}[t]
\centering
\caption{\small UE ($\mathrm{UE}_\text{MCQ}$, $\mathrm{UE}_\text{Open-QA}$), utility ($\mathrm{UT}_\text{MCQ}$ on MMLU/MathQA; $\mathrm{UT}_\text{Open-QA}$ on GSM8K/IFEval), and robustness ($\mathrm{Rob}_\text{FT}$, $\mathrm{Rob}_\text{ReL}$, $\mathrm{Rob}_\text{QT}$) of NPO and RMU on Qwen2.5-14B evaluated on WMDP.}
\vspace{-1.5mm}
\label{tab:qwen14b}
{\setlength{\tabcolsep}{8.8pt}%
\renewcommand{\arraystretch}{1.12}%
\resizebox{0.99\textwidth}{!}{%
\begin{tabular}{c | cc | cc cc | ccc c c}
\toprule[1pt]
\multirow{3}{*}{\textbf{Method}} &
\multicolumn{2}{c|}{\textbf{Unlearning Effectiveness ($\downarrow$)}} &
\multicolumn{4}{c|}{\textbf{Utility ($\uparrow$)}} &
\multicolumn{5}{c}{\textbf{Robustness ($\downarrow$)}} \\
\cline{2-3}\cline{4-7}\cline{8-12}

& \multirow{2}{*}{\textbf{$\mathrm{UE}_\text{MCQ}$}} &
  \multirow{2}{*}{\textbf{$\mathrm{UE}_\text{Open-QA}$}} &
\multicolumn{2}{c}{\textbf{$\mathrm{UT}_\text{MCQ}$ ($\uparrow$)}} &
\multicolumn{2}{c|}{\textbf{$\mathrm{UT}_\text{Open-QA}$ ($\uparrow$)}} &
\multicolumn{3}{c}{\textbf{$\mathrm{Rob}_\text{FT}$ ($\downarrow$)}} &
\multirow{2}{*}{\textbf{$\mathrm{Rob}_\text{ReL}$ ($\downarrow$)}} &
\multirow{2}{*}{\textbf{$\mathrm{Rob}_\text{QT}$ ($\downarrow$)}} \\
\cline{4-5}\cline{6-7}\cline{8-10}

& & &
\textbf{MMLU} & \textbf{MathQA} &
\textbf{GSM8K} & \textbf{IFEval} &
\textbf{GSM8K} & \textbf{MNLI} & \textbf{SST2} & & \\
\midrule

\rowcolor{gray!10}
NPO & 0.36 & 0.28 & 0.72 & 0.51 & 0.00 & 0.14 & 0.60 & 0.70 & 0.53 & 0.69 & 0.36 \\
RMU & 0.32 & 0.34 & 0.76 & 0.53 & 0.90 & 0.45 & 0.32 & 0.35 & 0.36 & 0.79 & 0.29 \\

\bottomrule[1pt]
\end{tabular}%
}}
\end{table*}

%% file: secs/quantization_analysis.tex
\noindent\textbf{Robustness to post-unlearning quantization.}
Quantization is a practically relevant form of weight perturbation \citep{zhang2024catastrophic}, but its effect cannot be judged from forget-set performance alone. Compression may reduce general model capability, in which case an apparent improvement in UE reflects inability to answer rather than selective removal of target knowledge. We therefore evaluate the joint UE--UT movement before and after quantization.

\begin{figure}[!b]
\centering
\includegraphics[width=0.90\columnwidth]{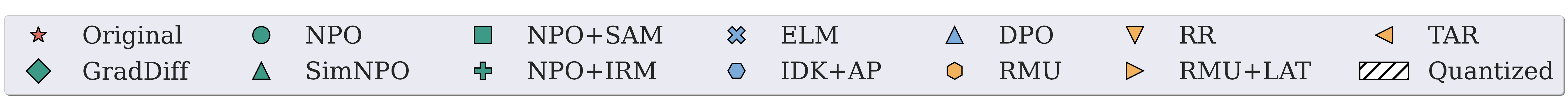} \\[-0.5mm]
\begin{tabular}{@{}c@{\hspace{1.5mm}}c@{}}
\includegraphics[width=0.42\columnwidth]{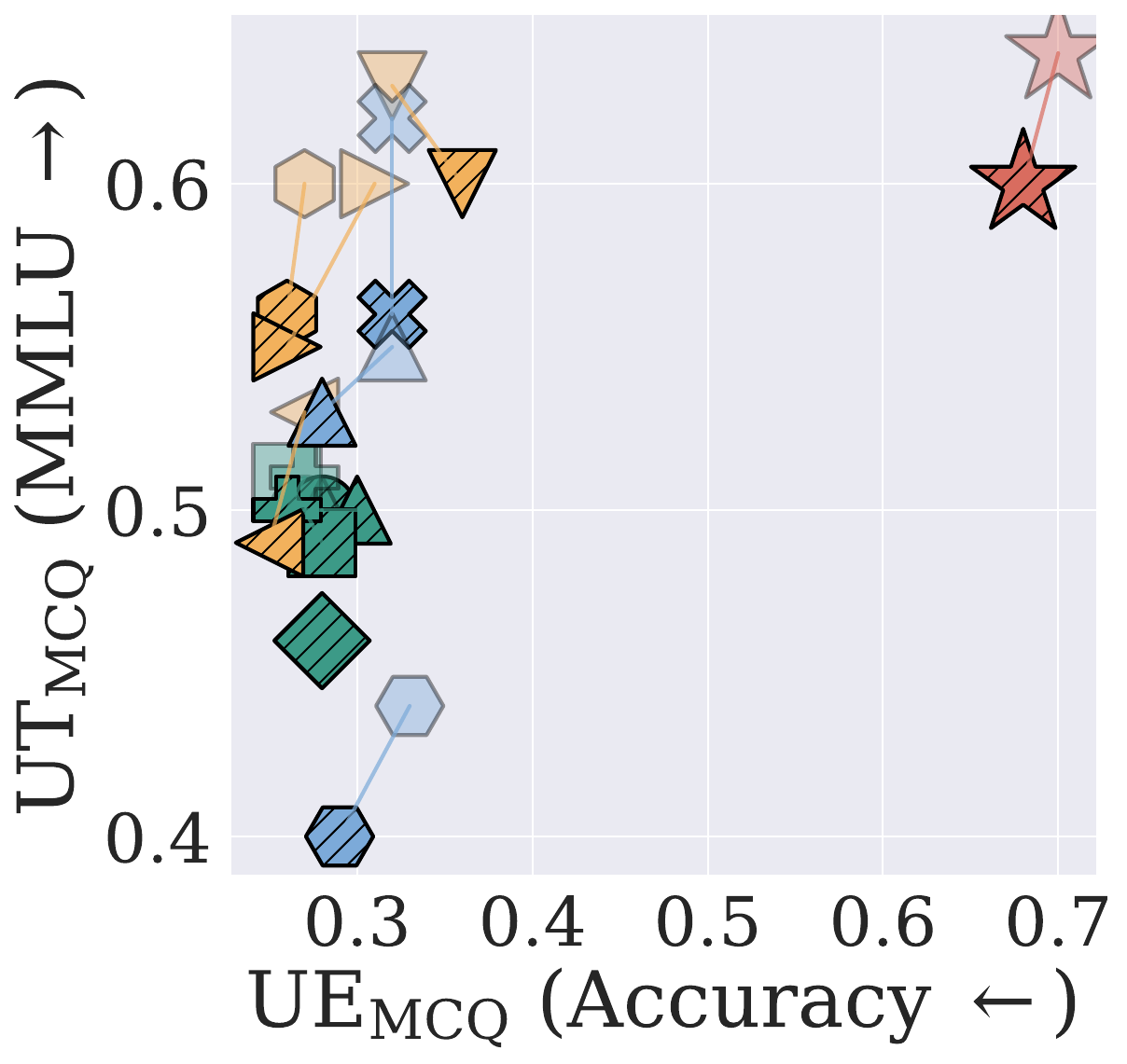} &
\includegraphics[width=0.42\columnwidth]{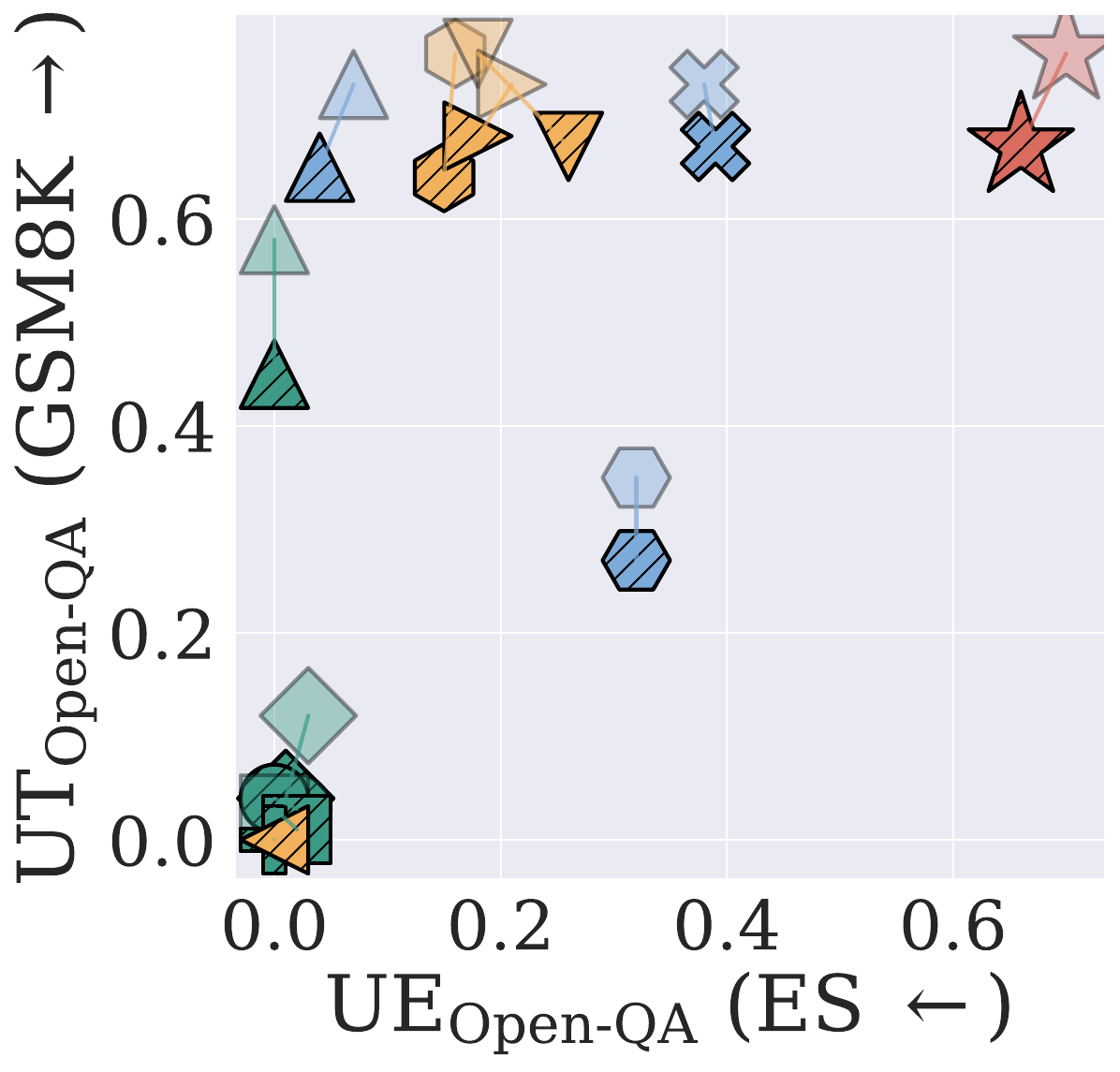} \\[-1mm]
\parbox{0.42\columnwidth}{\centering\footnotesize (a) $\mathrm{UE}_\text{MCQ}$ vs. $\mathrm{UT}_\text{MCQ}$} &
\parbox{0.42\columnwidth}{\centering\footnotesize (b) $\mathrm{UE}_\text{Open-QA}$ vs. $\mathrm{UT}_\text{Open-QA}$}
\end{tabular}
\vspace{-1.5mm}
\caption{\small Robustness to 4-bit quantization evaluated through the joint UE--UT trade-off. Lines connect each full-precision model to its quantized counterpart; hatched markers denote quantized models. Panel (a) uses WMDP accuracy and MMLU, while panel (b) uses WMDP ES and GSM8K.}
\Description{Two scatter plots showing joint changes in unlearning effectiveness and utility before and after 4-bit weight quantization under MCQ and Open-QA evaluations.}
\label{fig:rob_weight_quantization}
\end{figure}

\textbf{Fig.~\ref{fig:rob_weight_quantization}} shows that quantization can move UE and UT together, making a one-dimensional interpretation unreliable. \textbf{Table~\ref{tab:quantization_results}} further reveals that the effect depends on the unlearning setting. On MUSE, 4-bit NPO increases forget-set KnowMem from 2.10 to 30.30 (lower is better), while 4-bit RMU reduces retain-set KnowMem from 59.55 to 25.84; both changes undermine the desired UE--UT balance, although through different metrics. On WMDP, by contrast, MCQ UE is nearly unchanged across precision levels: NPO remains at 0.28 and RMU changes only from 0.27 to 0.26. MMLU utility is stable for NPO (0.46) and declines modestly for RMU (0.60 to 0.56). Thus, the observed quantization vulnerability is benchmark- and objective-dependent: data-centric content removal is markedly less stable here than WMDP knowledge removal, and robustness claims should report utility alongside forgetting.

%% file: secs/conclusion.tex
\section{Conclusion}

In this work, we establish a principled taxonomy of LLM unlearning for undesired model capability removal, organizing twelve representative unlearning methods into three families, {\MDiv}, {\MRep}, and {\MRej}, and providing a systematic lens for understanding their underlying mechanisms and relationships. Our analysis reveals that conventional MCQ-based evaluations of unlearning effectiveness offer an incomplete picture, and we introduce Open-QA as a complementary paradigm to better capture generative behavior and expose the strengths and limitations of different methods. Furthermore, we conduct a comprehensive robustness assessment, uncovering nuanced relationships among in-domain relearning, out-of-domain fine-tuning, and jailbreaking. Looking ahead, we plan to extend our study to data-wise unlearning and contrast it with model capability removal.

%% file: secs/limitations.tex
\noindent\textbf{Limitations.}
Our study has three main limitations. First, the taxonomy covers twelve representative stateful methods rather than all unlearning approaches; methods outside these families may exhibit different effectiveness--utility trade-offs. Second, input-level robustness centers on jailbreak prompts; in-context demonstrations, adaptive multi-turn attacks, and other prompting strategies may expose additional failure modes. Third, most comparisons rely on automatic metrics: combining MCQ and Open-QA reduces single-metric blind spots, yet human evaluation would provide complementary evidence on semantic quality, calibrated uncertainty, and user trust.

%% file: secs/ai_usage.tex
\noindent\textbf{Generative AI usage disclosure.}
GPT-5 was used during manuscript preparation for grammar correction and language refinement; it was not treated as an author and did not determine the research questions, experimental design, measurements, or conclusions.

%% file: secs/impact.tex
\section{Ethical Considerations}
\label{sec:ethics}

Reliable unlearning could reduce privacy leakage, limit reproduction of harmful or copyrighted material, and support compliance with data-removal obligations. Our taxonomy and multi-axis evaluation are intended to make these benefits easier to assess rather than to imply that a low score on one forget-set metric certifies safe removal.

Unlearning also creates risks. Excessive or poorly localized forgetting can degrade useful capabilities, while apparent forgetting may disappear under relearning, unrelated fine-tuning, quantization, or adversarial prompting. Conversely, a highly robust removal mechanism could be misused to suppress legitimate information or make model behavior less transparent. Deployment decisions should therefore consider the provenance and legitimacy of removal requests, report both effectiveness and collateral utility loss, test robustness under realistic post-deployment changes, and preserve auditable records of the intervention. We hope this work supports transparent unlearning practices that balance safety, utility, accountability, and robustness.